\documentclass{article}

\PassOptionsToPackage{numbers, compress}{natbib}
\usepackage[preprint]{neurips_2026}

\usepackage[utf8]{inputenc} % allow utf-8 input
\usepackage[T1]{fontenc}    % use 8-bit T1 fonts
\usepackage[colorlinks=true, allcolors=blue]{hyperref}       % hyperlinks
\usepackage{url}            % simple URL typesetting

\usepackage{booktabs}       % professional-quality tables
\usepackage{amsfonts}       % blackboard math symbols
\usepackage{nicefrac}       % compact symbols for 1/2, etc.
\usepackage{microtype}      % microtypography
\usepackage{xcolor}         % colors

\usepackage{graphicx}       % figures
\usepackage{amsmath}        % math
\DeclareMathSizes{6.8}{6.8}{5}{5}
\usepackage{multirow}       % tables with multi-row cells
\usepackage{threeparttable} % tables with notes
\usepackage{enumitem}       % compact lists / description environment
\usepackage{placeins}       % \FloatBarrier (used in the appendix)
\usepackage{listings}       % LLM prompt listings (appendix)
\usepackage{textcomp}       % \textcent etc.
\usepackage{comment}        % \begin{comment} blocks
\usepackage{fontawesome5}   % \faGithub icon

\newcommand{\sys}{\textsc{RealSWE}}
\newcommand{\sysb}{\textsc{RealSWE-bench}}
\newcommand{\sysf}{\textsc{RealSWE-framework}}

\title{RealSWE: A Compositional Evaluation \\ of Coding Agents under Realistic User Requests}

\author{%
  Gyuhyeong Kim \quad Hyojung Gwon \quad Jeonghyeon Kim \quad
  Kyuhong Shim \quad Sunjae Lee\thanks{Corresponding author.} \\
  Sungkyunkwan University \\
  \texttt{\{gyuhyeong, gywndgywnd12, jeonghyeon12, khshim, sunjae.lee\}@skku.edu}
}

\begin{document}

\maketitle

\vspace{-1.6em}
\begin{center}
\makebox[0pt][r]{\faGithub~\href{https://github.com/gyuhyeong-k/RealSWE}{GitHub}\hspace{1em}}%
\makebox[0pt][l]{\hspace{1em}\raisebox{-0.17em}{\includegraphics[trim=23 30.25 25 30, clip, height=1.12em]{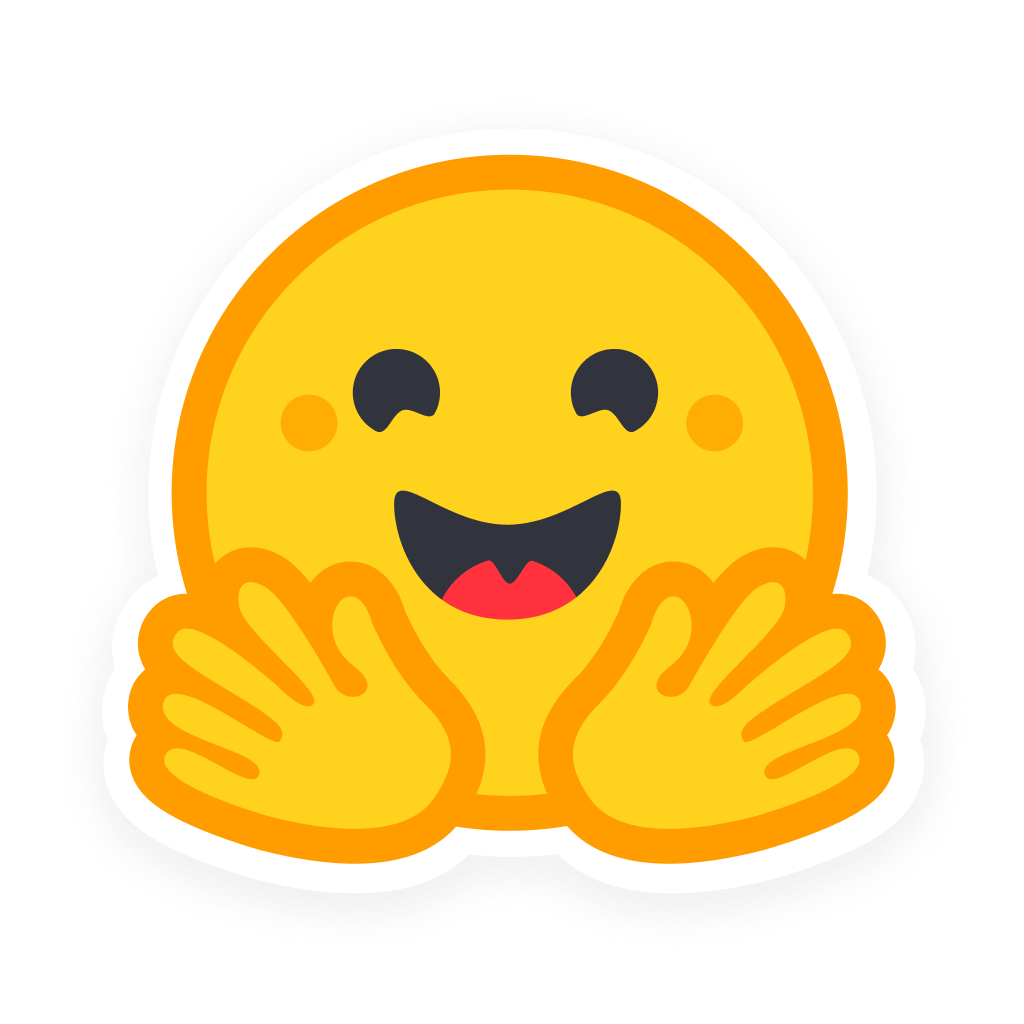}}~\href{https://huggingface.co/datasets/gyuhyeong-k/RealSWE-bench}{HuggingFace}}
\end{center}
\vspace{0.7em}

% ============================================================
% Body
% ============================================================
\begin{abstract}
Coding agents are now commonly evaluated on the \textsc{SWE-bench} family of benchmarks, whose tasks are built from curated GitHub issues---long, structured, and information-rich. Real user requests, however, are typically far shorter and less structured. To characterize this gap, we define a six-category information taxonomy and four dimensions of linguistic style, and apply them to real user prompts from \textsc{SWE-chat} and problem statements from \textsc{SWE-bench Verified} and \textsc{Pro}. We find that requests carrying only a problem statement, alone or with limited additional context, account for 88\% of real prompts but just 7\% of benchmark problems. Furthermore, 87\% of real prompts are casually written whereas 94\% of benchmark problems are formal. Guided by these observations, we introduce \sys{}, 381 \textit{multi-variant task families} derived from \textsc{SWE-bench Verified} and \textsc{Pro}. Variants within each family share the same underlying task and gold patch while differing only in information composition and linguistic style. Evaluating seven contemporary LLMs with \sys{}, we find that \textit{i)} realistic inputs reduce resolution rates by 6.4 pp on average and can change model rankings. Controlled analysis further shows that \textit{ii)} including Desired Behavior and Motivation significantly affects performance, whereas Environment Information and Reproduction Steps merely add tokens without measurable benefit; \textit{iii)} linguistic style has only small, model-dependent effects. These findings provide actionable guidance for users and agents: explicitly stating the desired behavior and motivation---which most real prompts omit---substantially improves the LLM's software engineering performance.
\end{abstract}
\section{Introduction}

\begin{figure}[t]
  \centering
  \includegraphics[width=0.7\linewidth]{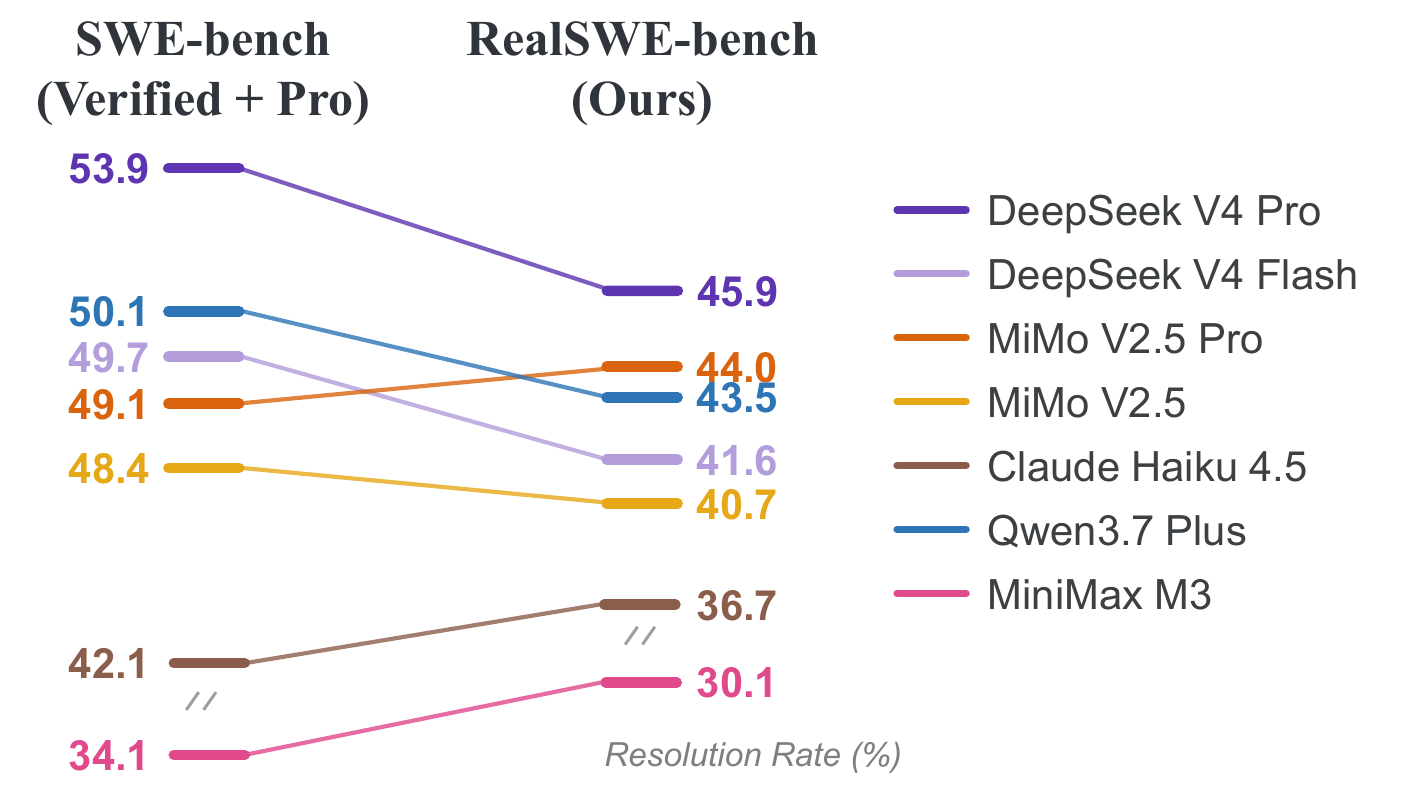}
  \caption{Resolution rates on the original \textsc{SWE-bench} tasks and \sysb{}. All models decline under realistic inputs.}
  \label{fig:swebench_vs_realswe}
\end{figure}

Large language models (LLMs) are rapidly transforming software engineering, advancing beyond function-level code generation toward coding agents that autonomously resolve repository-level issues. Their progress is commonly measured using \textsc{SWE-bench} families~\citep{jimenez2024swebench, openai2024swebenchverified, zan2025multiswebench, rashid2025swepolybench, deng2025swebenchproaiagents, zhang2025swebenchlive, adamenko2025swemera}, which harvest executable tasks from GitHub issues and their corresponding fixes (e.g., commits and pull requests) in well-known open-source GitHub repositories. Today, leaderboard scores on these benchmarks serve as the de facto standard for comparing LLMs' coding capabilities.

Yet a growing body of evidence indicates that the inputs these benchmarks feed to agents are \textit{far} from what agents receive in practice. Curated GitHub issues are typically detailed, well-structured, and long, whereas everyday user requests are short, informal, and sparse~\citep{cursor2026cursorbench, garg2026savingswebench, baumann2026swechat}. For example, a GitHub issue may describe a failure, provide reproduction steps and environment information, specify the desired behavior, and even suggest a solution. A user may express the same intent simply as \textit{``this crashes on empty input---fix it.''} Although the underlying task is identical, the latter provides far less evidence, requiring the agent to infer missing requirements from the repository and context. Benchmark performance may therefore depend not only on task difficulty, but also on how the task is communicated.

Recent work has begun to close this gap with new \textit{realistic} benchmarks. \textsc{CursorBench}~\citep{cursor2026cursorbench} evaluates prompts drawn from real coding sessions and others synthesize underspecified tasks~\citep{chen2026nlperturbator, zhuo2025bigcodebench, wu2025humanevalcomm} or mutate GitHub issues~\citep{garg2026savingswebench, sweinteract2026, ambigswe2026} to simulate user inputs. However, these efforts fall short in two ways. Benchmarks grounded in genuine user data, such as \textsc{CursorBench}, are closed-source and unavailable for independent use or inspection. Open alternatives approximate realism heuristically, for example by truncating problem statements or injecting ambiguity, without systematic analysis or empirical grounding in actual user data. Consequently, while they reveal that a benchmark--reality gap \textit{exists}, they fail to provide insight into \textit{what} causes it or \textit{how} it should be measured.

In this paper we address this gap through a systematic, data-grounded approach. We begin by analyzing user prompts from \textsc{SWE-chat}, a large-scale dataset of real interactions between developers and coding agents~\citep{baumann2026swechat}. We characterize each request along two orthogonal axes: the \textit{information} it conveys (Table~\ref{tab:schema}) and the \textit{language} in which it is conveyed (Figure~\ref{fig:Overall}).
Our analysis reveals a substantial mismatch between real user requests and benchmark problems on both axes. Real requests are information-sparse: prompts consisting only of a problem statement (i.e., which bug to fix or which feature to implement) alone or with limited additional context (e.g., source URLs, dates) account for 88\% of user prompts, compared with only 7\% of tasks in \textsc{SWE-bench Verified} and \textsc{SWE-bench Pro}. They also differ linguistically: 87\% of user prompts use a casual tone and 51\% contain imperative sentences, whereas 94\% of benchmark problems use a formal tone and 89\% rely on declarative sentences. These differences provide an empirical basis for building realistic coding-agent evaluation.

Guided by these observations, we introduce \sys{}, an open benchmark and configurable evaluation framework grounded in the characteristics of real user inputs. It contains 381 multi-variant task families derived from \textsc{SWE-bench Verified} and \textsc{Pro}. Each family shares the same underlying task while varying information composition and linguistic style. We release it as \sysb{}, a fixed evaluation set matching the distributions observed in \textsc{SWE-chat}, and \sysf{}, which exposes the full variant suite for custom configurations and controlled ablations.

We evaluate seven contemporary LLMs under \sys{}. The results reveal substantial discrepancies between performance on original \textsc{SWE-bench}-style problems and realistic user inputs: resolution rates drop by 6.4 percentage points (pp) on average, and the gap between stronger and weaker models narrows (Figure~\ref{fig:swebench_vs_realswe}). More importantly, our controlled analysis under \sysf{} reveals a sharply uneven value of information. For bug fixes, the presence of Desired Behavior strongly affects success ($+$8 pp, 17\% relative), while Reproduction Steps and Environment Information add input tokens with no measurable benefit; for feature requests, adding Motivation improves performance by up to $+$7 pp. Linguistic style changes, in contrast, produce only small, model-dependent effects. These findings expose an actionable mismatch between what users provide and what agents need: Desired Behavior of a bug fix and Motivation behind the new feature are among the most valuable signals an agent can receive, yet only 5\% of real user prompts state it. Explicitly stating these can thus substantially improve the performance users experience from coding agents.

In summary, our contributions are threefold:
\begin{itemize}[leftmargin=*]
    \item \textbf{Empirical characterization of real SWE requests.}
    Using a structured information taxonomy and linguistic dimensions, we analyze how real user requests in \textsc{SWE-chat} are composed, and quantify how their information composition and linguistic style differ from those of \textsc{SWE-bench Verified} and \textsc{SWE-bench Pro}.
    
    \item \textbf{A data-grounded benchmark and configurable framework.}
    Guided by this analysis, we construct 381 multi-variant task families and release them in two forms: \sysb{}, an open-source benchmark reflecting the empirical distribution of real user inputs, and \sysf{}, a configurable framework that allows researchers to customize the information composition and linguistic style of the evaluation.
    
    \item \textbf{Controlled evaluation and actionable findings.}
    Evaluating seven LLMs, we \textit{i)} quantify the benchmark--reality gap, \textit{ii)} identify Desired Behavior and Motivation as key signals for software engineering tasks, and \textit{iii)} distill actionable guidance for users of coding agents.
\end{itemize}
\section{Related Work}

\begin{table}[t]
  \caption{Information taxonomy and field abbreviations for bug-fix and feature-request prompts.}
  \label{tab:schema}
  \small
  \centering
  \begin{tabular}{@{}l @{\hspace{0.6em}} l @{\hspace{2em}} l @{\hspace{0.6em}} l@{}}
    \toprule
    \multicolumn{2}{@{}l}{\textbf{Bug fix}} & \multicolumn{2}{@{}l}{\textbf{Feature request}} \\
    \midrule
    \texttt{[P]} & \begin{tabular}[t]{@{}l@{}}Problem Statement\\{\itshape\color{black!60}(Observed Failure)}\end{tabular} & \texttt{[P]} & \begin{tabular}[t]{@{}l@{}}Problem Statement\\{\itshape\color{black!60}(Requested Feature)}\end{tabular} \\
    \addlinespace[2pt]
    \texttt{[D]} & Desired Behavior & \texttt{[M]} & Motivation \\
    \addlinespace[2pt]
    \texttt{[R]} & Reproduction Steps & \texttt{[A]} & Additional Information \\
    \addlinespace[2pt]
    \texttt{[E]} & Environment Information & & \\
    \addlinespace[2pt]
    \texttt{[A]} & Additional Information & & \\
    \bottomrule
  \end{tabular}
\end{table}

\paragraph{Coding agent benchmarks.}
\textsc{SWE-bench} established repository-level issue resolution as a standard evaluation setting by pairing real GitHub issues with executable tests~\citep{jimenez2024swebench}. Subsequent benchmarks have improved reliability and expanded language, repository, modality, task, and complexity coverage~\citep{openai2024swebenchverified,zan2025multiswebench,rashid2025swepolybench,yang2024swebenchmultimodalaisystems,deng2025swebenchproaiagents,li2025feabenchbenchmarkevaluatingrepositorylevel}. However, these benchmarks generally associate each executable task with a single canonical issue description. They therefore broaden and strengthen the underlying software engineering tasks, but do not examine how performance changes when the same task is communicated with different information or linguistic forms.

\paragraph{Realistic and communication-aware evaluation.}
Recent work has begun to incorporate realistic tasks using naturally occurring developer requests or transformed benchmark inputs. Benchmarks built from real coding sessions provide authentic user inputs, but communication varies together with the task, repository, and difficulty, making its independent effect difficult to isolate~\citep{cursor2026cursorbench,jha2026reapautomaticcurationcoding}. Observational datasets such as \textsc{SWE-chat} further characterize how developers communicate with coding agents in practice~\citep{baumann2026swechat}. Other work introduces underspecified or interactive variants to study ambiguity and clarification behavior~\citep{ambigswe2026,edwards2026askassumeuncertaintyawareclarificationseeking,king2026dialogueswebenchbenchmarkdialoguedriven}.

Most closely related to our work, \textsc{Saving SWE-Bench} uses patterns observed in real developer interactions to transform existing repository tasks into user-style inputs~\citep{garg2026savingswebench}. However, its transformations jointly alter multiple properties of the task specification, making it difficult to attribute performance changes to particular information components or linguistic properties. In contrast, \sys{} represents each task as a multi-variant family grounded in the information compositions and linguistic dimensions observed in real requests. This design independently controls the information content and linguistic style of a fixed repository task, enabling per-field ablations, arbitrary compositions, and distribution-matched evaluation.
\section{Method}

\begin{figure}[t]
  \centering
  \includegraphics[width=\linewidth]{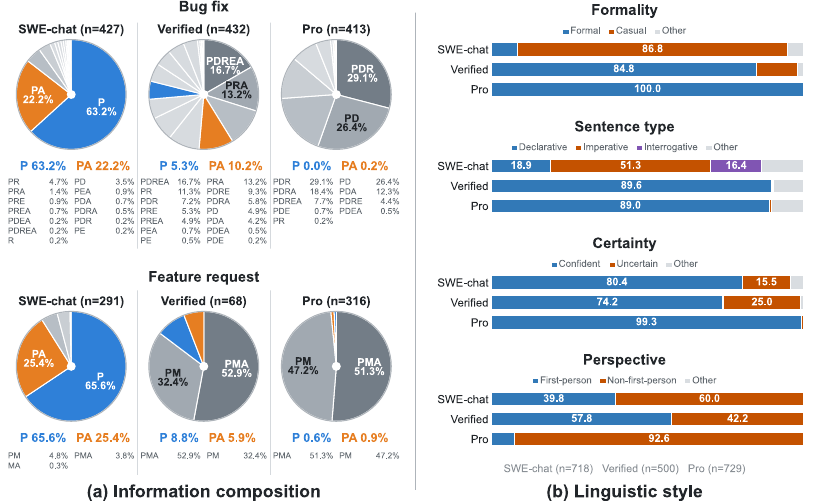}
  \caption{Distributions of (a) information composition and (b) linguistic style across \textsc{SWE-chat}, \textsc{SWE-bench Verified}, and \textsc{SWE-bench Pro}. See Appendix~\ref{app:mismatch} for detailed results.}
  \label{fig:Overall}
\end{figure}

\begin{figure}[t]
  \centering
  \includegraphics[width=\linewidth]{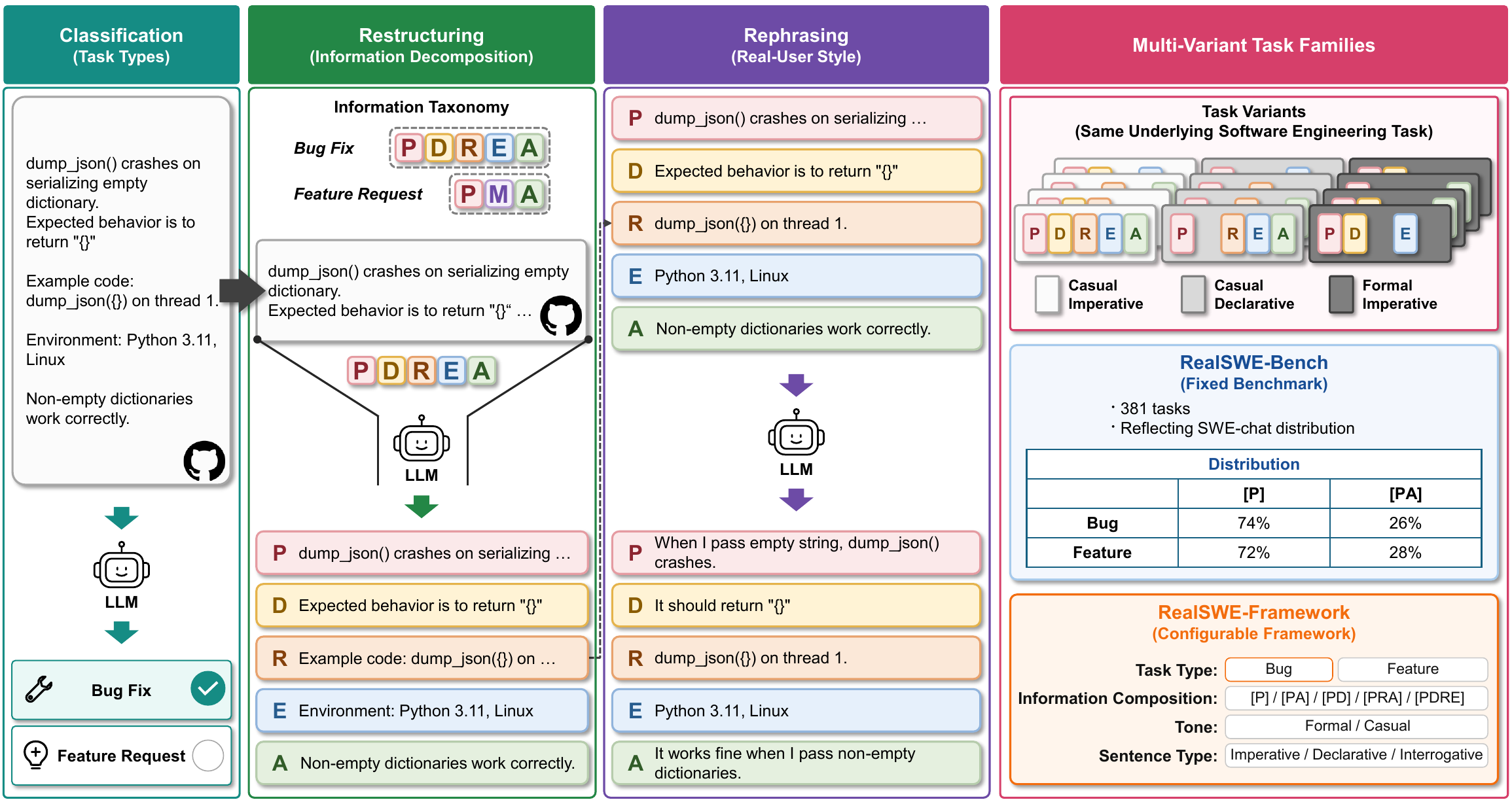}
  \caption{Overview of the \sys{} construction pipeline.}
  \label{fig:Our_Overview}
\end{figure}

We introduce \sys{} (Figure~\ref{fig:Our_Overview}), a benchmark and configurable framework for evaluating the same software engineering task under systematically varied task specifications, including \sysb{}, a fixed configuration that reflects the characteristics of real user requests. We construct \sys{} in four stages. First, drawing on established software engineering practices and prior literature, we define task-specific \textit{information taxonomies} and \textit{linguistic dimensions} for characterizing software engineering requests (\S\ref{sec:characterizing}). Second, we apply this scheme to real user requests from \textsc{SWE-chat} and problem statements from \textsc{SWE-bench Verified} and \textsc{SWE-bench Pro}, quantifying their differences in information composition and linguistic style (\S\ref{sec:mismatch}). Third, guided by these measurements, we transform the benchmark problems into \textit{multi-variant task families} that represent the same software engineering task under different information compositions and linguistic styles (\S\ref{sec:constructing}). Fourth, we validate each LLM-driven stage of the construction pipeline against human judgment (\S\ref{sec:validation}). Finally, we release a fixed benchmark reflecting dominant real-user input patterns (\sysb{}) and a configurable framework exposing all task variants (\sysf{}) (\S\ref{sec:benchmark}).

\subsection{Characterizing SWE Requests}
\label{sec:characterizing}

\paragraph{Data.} We use two complementary data sources. For real user inputs, we analyze \textsc{SWE-chat}~\citep{baumann2026swechat}, a public dataset containing more than 6,000 real developer--agent sessions. Because \textsc{SWE-bench} typically evaluates an agent from a single problem statement without further interaction, we retain only the first user request from each session. We further remove prompts that do not express an actionable SWE task, including conversational dialogue, inputs generated by other tools, and pasted LLM outputs, leaving 718 user-authored prompts (Appendix~\ref{app:swechat}).

We compare these real requests against problem statements from \textsc{SWE-bench Verified}~\citep{openai2024swebenchverified} and \textsc{Pro}~\citep{deng2025swebenchproaiagents}. We choose these two benchmarks because they represent widely used repository-level coding-agent evaluation. \textsc{SWE-bench Verified} is a human-validated benchmark for standardized model comparison. \textsc{SWE-bench Pro} extends this setting to more difficult, longer-horizon tasks drawn from larger and more complex repositories.

\paragraph{Task types.}
Following common practice in issue-tracking systems~\citep{2486788.2486840}, we categorize each request as either a \textit{bug fix}---existing behavior is incorrect and needs a fix---or a \textit{feature request}---the user asks for new or changed functionality. These two types cover nearly all requests: only 2 of the 1,231 problem statements in \textsc{SWE-bench Verified} and \textsc{Pro} fall into neither, and we exclude them before the rest of the pipeline (Table~\ref{tab:app-classification}).

\paragraph{Information taxonomy.}
For each task type, we define an \textit{information taxonomy}: the set of information types that a request may contain. The taxonomy is grounded in GitHub's default issue templates~\citep{githubIssueTemplates, sulun2024issuetemplates} (Table~\ref{tab:schema}). \texttt{[A]} captures information outside the other categories, such as source URLs, issue-author metadata, and dates. A request is then described by the \textit{set} of information types it contains---e.g., \texttt{[P]}, \texttt{[PA]}, or \texttt{[PDR]}.

\paragraph{Linguistic properties.}
To characterize how a request is written, we categorize its linguistic style along four dimensions: \textit{Formality}, \textit{Sentence type}, \textit{Certainty}, and \textit{Perspective} (Figure~\ref{fig:Overall}). Adapted from~\citep{truong-etal-2025-persona}, these dimensions are chosen to separate GitHub issue-style prose from the conversational language of coding-agent chats.

\subsection{Measuring the Benchmark--Reality Mismatch}
\label{sec:mismatch}

\paragraph{Annotation procedure.}
We apply the information taxonomy and linguistic dimensions to the 718 user-authored \textsc{SWE-chat} requests and to the 1,229 problem statements from \textsc{SWE-bench Verified} and \textsc{Pro}. Using an LLM-assisted pipeline, GPT-5.4~\citep{openai2026gpt54} identifies each piece of information in a prompt, assigns it to a taxonomy category, and classifies the prompt's style along the four dimensions.

\paragraph{Results: the mismatch.}
Figure~\ref{fig:Overall} summarizes the resulting distributions. Most notably, the two sources diverge sharply in information composition. Real requests are compositionally sparse: \texttt{[P]} and \texttt{[PA]} alone account for 88\% of prompts (85.5\% of bug-fix and 91.1\% of feature requests), indicating that in real-world practice, users often rely on simple, underspecified requests such as \textit{``Server crashes on empty input, fix it.''} or \textit{``Implement new feature that does ...''} In comparison, problems in \textsc{SWE-bench Verified} and \textsc{Pro} are information-rich, with only 7\% consisting of \texttt{[P]}/\texttt{[PA]} (8.0\% for bug fixes, 3.9\% for feature requests). They often include additional fields such as Reproduction Steps, Environment Information, and other contextual details that real users rarely provide. This gap likely leads benchmarks to overestimate LLMs' coding performance in real-world settings.

The linguistic dimensions exhibit a less uniform pattern (Figure~\ref{fig:Overall}(b)). \textsc{SWE-chat} and the \textsc{SWE-bench} datasets differ most strongly in formality and sentence type: 86.8\% of real user requests are casual and 51.3\% are imperative. In contrast, 84.8\% and 100\% of prompts in \textsc{SWE-bench Verified} and \textsc{Pro}, respectively, are formal, and approximately 89\% of prompts in both benchmarks are declarative. This suggests that \textsc{SWE-bench} prompts resemble polished issue reports rather than conversational user requests. Certainty and perspective show no consistent separation between real requests and benchmark prompts. These distributions serve as empirical targets for realistic evaluation and directly guide the construction described below.

\subsection{Constructing Multi-Variant Task Families}
\label{sec:constructing}

Guided by the observed distributions, we transform each benchmark problem from \textsc{SWE-bench Verified} and \textsc{Pro} into a \textit{multi-variant task family} through a three-step, LLM-driven pipeline (task-type classification, information decomposition, and real-user-style rephrasing). The complete prompts for all three steps are provided in Appendix~\ref{app:prompts}.

\paragraph{Decomposing problem statements.}
First, we leverage {GPT-5.4} to classify each task as a bug fix or feature request. Before decomposition, we also strip scaffolding inherited from GitHub issue templates, such as Markdown headings and HTML tags, while preserving all user-written text (Appendix~\ref{app:pipeline:decomposition}). Then, we segment each original problem statement into sentences and assign each sentence to an information taxonomy category. We redistribute the original text across these fields without rewriting it so that decomposition does not alter the information it carries.

\paragraph{Rephrasing in real-user style.}
We rewrite the restructured problems into the style observed in \textsc{SWE-chat} using {GPT-5.4}, conditioning on the majority category of each linguistic dimension measured in \S\ref{sec:mismatch}. This process changes only the manner of expression while preserving technical content, such as code blocks, error messages, tracebacks, and file paths (Appendix~\ref{app:pipeline:rephrasing}).

\paragraph{Collecting task families with configurable variants.}
\sys{} represents each software engineering problem using arbitrary combinations of information categories. Supporting such configurations requires every taxonomy field to be present in the source problem statement. Of the 1,229 problems in \textsc{SWE-bench Verified} and \textsc{SWE-bench Pro}, 403 contain all required information categories, forming our initial candidate set (Appendix~\ref{app:pipeline:decomposition}).

\subsection{Validation \& Quality Control}
\label{sec:validation}

We assess both the reliability of the construction pipeline and the quality of the resulting task families through human validation. For each LLM-driven stage, two annotators independently evaluate 100 sampled instances and resolve disagreements by consensus to establish human ground truth. Appendix~\ref{app:validation} reports the full rubrics, inter-annotator agreement, and validation results.

\paragraph{Pipeline validation and task quality.}
For task-type classification, we directly measure accuracy against the human ground truth; the classification pipeline achieves an accuracy of 0.95. For field decomposition and linguistic rephrasing, we use GPT-5.6 Terra~\citep{openai2026gpt56} as an LLM judge to audit all candidate tasks using the same three-point rubrics as the human annotators. First, we assess the quality of the task set by verifying whether the field-decomposition pipeline assigns each content unit to the correct taxonomy field and whether the resulting task specification remains complete and coherent. Second, we assess the realism of the transformed requests by verifying whether the rephrasing pipeline produces the target real-user style while preserving the original information, meaning, implementation intent, and technical literals.

The LLM judge shows high agreement with human judgment (decomposition: accuracy 0.97, macro-\(F_1\) 0.83; rephrasing: accuracy 0.99, macro-\(F_1\) 0.75). We exclude candidates receiving the lowest score on any critical criterion, removing 22 of the 403 candidates and leaving \textbf{381 task families (192 bug fixes and 189 feature requests)}.

\paragraph{Selection bias.}
Our selection process reduces the original pool of 1,229 tasks to 381. To assess potential selection bias, we compare the selected 381 tasks with the excluded 848 tasks in terms of resolution rate, patch size, and repository distribution. We find that the selected task families are not easier, smaller, or concentrated in particular repositories. Appendix~\ref{app:validation:selection} reports the complete comparison.

\subsection{\sys{} Benchmark and Framework}
\label{sec:benchmark}

To reflect the information distribution observed in \textsc{SWE-chat} (\S\ref{sec:mismatch}), \sysb{} samples one variant per task family in the following proportions: \texttt{[P]} 74\% and \texttt{[PA]} 26\% for bug fixes, and \texttt{[P]} 72\% and \texttt{[PA]} 28\% for feature requests. This yields 142 \texttt{[P]} and 50 \texttt{[PA]} bug-fix tasks, and 136 \texttt{[P]} and 53 \texttt{[PA]} feature-request tasks (total 381 tasks).

\begin{figure}[t]
  \centering
  \includegraphics[width=0.7\linewidth]{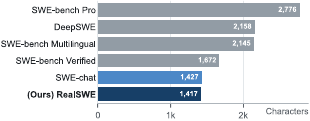}
  \caption{Mean task-description length across coding benchmarks and the real user prompts in \textsc{SWE-chat}. External documents (e.g., \texttt{AGENTS.md} and documentation) are excluded. \sysb{} closely matches \textsc{SWE-chat}, whereas conventional benchmarks are substantially longer.}
  \label{fig:prompt_length}
\end{figure}

The resulting benchmark has an average task-description length of 1,417 characters, closely matching the 1,427 average observed in \textsc{SWE-chat} and well below the 1,672--2,776 characters of conventional SWE benchmarks (\textsc{SWE-bench Verified}, \textsc{Multilingual}, \textsc{Pro}, and \textsc{DeepSWE}; Figure~\ref{fig:prompt_length}). Given that prior work identifies description length as a key benchmark--reality distinction~\citep{cursor2026cursorbench}, this alignment corroborates the realism of \sysb{}.

\sysf{} exposes all 381 task families through a configuration interface. Researchers specify a task type, information composition, and linguistic style---for example, \textit{bug fixes containing only \texttt{[P]} and \texttt{[D]} in a casual style}---and the framework assembles the corresponding dataset on demand.
\section{Experiments}
\label{sec:experiments}

\begin{table}[b]
\caption{Resolution rate (\%) on the original problem statements and \sysb{}. $\Delta$ denotes \sysb{} minus Original; the largest drop in each $\Delta$ column is in bold.}
\label{tab:main-results}
\centering
\footnotesize
\setlength{\tabcolsep}{4pt}
\begin{tabular*}{\textwidth}{@{}l@{\hspace{1em}\extracolsep{\fill}} ccc ccc ccc@{}}
\toprule
& \multicolumn{3}{c}{\textbf{All} ($n$=381)} & \multicolumn{3}{c}{\textbf{Bug fix} ($n$=192)} & \multicolumn{3}{c}{\textbf{Feature request} ($n$=189)} \\
\cmidrule(l{2pt}r{2pt}){2-4} \cmidrule(l{2pt}r{2pt}){5-7} \cmidrule(l{2pt}r{2pt}){8-10}
Model & Original & \textsc{RealSWE} & $\Delta$ & Original & \textsc{RealSWE} & $\Delta$ & Original & \textsc{RealSWE} & $\Delta$ \\
\midrule
DeepSeek V4 Pro & 53.9$_{\pm0.6}$ & 45.9$_{\pm2.0}$ & \textbf{\boldmath$-$8.0} & 59.4$_{\pm1.4}$ & 49.0$_{\pm2.3}$ & $-$10.4 & 48.3$_{\pm0.3}$ & 42.9$_{\pm2.3}$ & $-$5.5 \\
DeepSeek V4 Flash & 49.7$_{\pm1.7}$ & 41.6$_{\pm0.9}$ & \textbf{\boldmath$-$8.0} & 56.2$_{\pm1.0}$ & 44.4$_{\pm1.8}$ & \textbf{\boldmath$-$11.8} & 43.0$_{\pm2.5}$ & 38.8$_{\pm0.3}$ & $-$4.2 \\
MiMo V2.5 Pro & 49.1$_{\pm1.2}$ & 44.0$_{\pm1.0}$ & $-$5.1 & 54.9$_{\pm1.6}$ & 47.2$_{\pm2.1}$ & $-$7.6 & 43.2$_{\pm2.1}$ & 40.7$_{\pm1.1}$ & $-$2.5 \\
MiMo V2.5 & 48.4$_{\pm1.4}$ & 40.7$_{\pm1.5}$ & $-$7.7 & 54.5$_{\pm1.3}$ & 45.1$_{\pm1.2}$ & $-$9.4 & 42.2$_{\pm1.6}$ & 36.2$_{\pm2.1}$ & \textbf{\boldmath$-$6.0} \\
Claude Haiku 4.5 & 42.1$_{\pm1.1}$ & 36.7$_{\pm2.9}$ & $-$5.4 & 50.3$_{\pm1.6}$ & 40.1$_{\pm2.4}$ & $-$10.2 & 33.7$_{\pm1.9}$ & 33.2$_{\pm3.4}$ & $-$0.5 \\
Qwen3.7 Plus & 50.1$_{\pm0.5}$ & 43.5$_{\pm0.8}$ & $-$6.6 & 54.7$_{\pm1.8}$ & 45.5$_{\pm2.2}$ & $-$9.2 & 45.5$_{\pm2.4}$ & 41.4$_{\pm1.3}$ & $-$4.1 \\
MiniMax M3 & 34.1$_{\pm0.5}$ & 30.1$_{\pm1.7}$ & $-$4.0 & 40.5$_{\pm0.3}$ & 35.2$_{\pm0.3}$ & $-$5.2 & 27.7$_{\pm1.3}$ & 24.9$_{\pm3.2}$ & $-$2.8 \\
\bottomrule
\end{tabular*}
\end{table}

\begin{table}[t]
\caption{Average per-task cost and steps on the original problem statements and \sysb{}.}
\label{tab:cost-steps}
\centering
\small
\setlength{\tabcolsep}{8pt}
\begin{tabular}{@{}l @{\hspace{1.5em}} cc @{\hspace{1.5em}} cc@{}}
\toprule
& \multicolumn{2}{c}{\textbf{Cost} (\textcent)} & \multicolumn{2}{c}{\textbf{Steps}} \\
\cmidrule(l{2pt}r{2pt}){2-3} \cmidrule(l{2pt}r{2pt}){4-5}
Model & Original & \textsc{RealSWE} & Original & \textsc{RealSWE} \\
\midrule
DeepSeek V4 Pro   & $\phantom{0}3.02_{\pm0.02}$ & $\phantom{0}3.16_{\pm0.08}$ & $42.6_{\pm0.4}$ & $44.2_{\pm0.3}$ \\
DeepSeek V4 Flash & $\phantom{0}1.27_{\pm0.01}$ & $\phantom{0}1.33_{\pm0.00}$ & $47.7_{\pm0.4}$ & $49.0_{\pm0.2}$ \\
MiMo V2.5 Pro     & $\phantom{0}5.55_{\pm0.07}$ & $\phantom{0}6.52_{\pm0.18}$ & $47.9_{\pm0.4}$ & $48.6_{\pm0.5}$ \\
MiMo V2.5         & $\phantom{0}2.24_{\pm0.06}$ & $\phantom{0}2.31_{\pm0.04}$ & $47.3_{\pm1.1}$ & $47.5_{\pm0.5}$ \\
Claude Haiku 4.5  & $33.89_{\pm0.21}$ & $33.23_{\pm0.45}$ & $66.9_{\pm0.4}$ & $66.7_{\pm0.3}$ \\
Qwen3.7 Plus      & $15.26_{\pm0.67}$ & $16.12_{\pm0.23}$ & $56.2_{\pm0.7}$ & $56.8_{\pm0.4}$ \\
MiniMax M3        & $17.78_{\pm0.17}$ & $18.05_{\pm0.14}$ & $82.2_{\pm0.7}$ & $83.5_{\pm0.4}$ \\
\bottomrule
\end{tabular}
\end{table}

Using \sys{} (both benchmark and framework), we conduct a controlled evaluation of how realistic user inputs affect coding agents, organized around three research questions:
\begin{description}[leftmargin=!, labelwidth=2em, labelsep=0.3em]
    \item[\textbf{RQ1.}]
    \textit{
    How does agent performance change
    when benchmark problems are replaced with realistic user inputs?}

    \item[\textbf{RQ2.}]
    \textit{
    Does the linguistic style of a request alone affect coding-agent performance?}

    \item[\textbf{RQ3.}]
    \textit{
    Which information fields most affect task resolution, and how does their value differ?}
\end{description}

\subsection{Experimental Setup}
\label{sec:setup}

\paragraph{Models.}
We evaluate seven LLMs with varying sizes and families: {DeepSeek V4 Pro} and {DeepSeek V4 Flash}~\citep{deepseek2026v4}, {MiMo V2.5 Pro} and {MiMo V2.5}~\citep{xiaomi2026mimo}, {Claude Haiku 4.5}~\citep{anthropic2025haiku45}, {Qwen3.7 Plus}~\citep{qwen2026qwen37}, and {MiniMax M3}~\citep{minimax2026m3}. The set includes both open-weight and commercial models and spans a broad range of baseline performance. We enable reasoning for every model to match contemporary coding-agent use. Exact model snapshots, providers, and inference settings are reported in Appendix~\ref{app:experimental-details}.

\paragraph{Agent scaffold and execution environment.}
Following common practice in SWE benchmark evaluation, we run every model with the same mini-SWE-agent v2 scaffold~\citep{yang2024sweagent}. Its minimal Bash-only interface provides a consistent agent setup across model providers without relying on provider-specific tool-calls. Each task runs in the execution container released with its source benchmark, with a maximum of 100 agent steps and no cost limit. We run every model--condition pair three times and report the mean and standard deviation across runs. Details of the statistical analysis are provided in Appendix~\ref{app:results}.

\subsection{Main Results: The Benchmark--Reality Gap}
\label{sec:gap}

Table~\ref{tab:main-results} compares performance on the original problem statements with performance on \sysb{}. 
As a result, all seven models resolve fewer tasks under \sysb{}. The resolution rate decreases by 6.4 pp on average, with absolute drops ranging from 4.0 pp for {MiniMax M3} (34.1\% to 30.1\%) to 8.0 pp for {DeepSeek V4 Pro} (53.9\% to 45.9\%). Relative to each model's original score, the declines fall within a range of 10.3--16.2\%, with a mean of 13.6\%. The degradation is not specific to one model family; all models lose a similar fraction of their baseline performance. Conventional leaderboard scores should therefore be interpreted as optimistic estimates of performance under realistic, sparse requests.

The performance drop is accompanied by a modest increase in computation (Table~\ref{tab:cost-steps}). Six of the seven models incur higher cost under \sysb{} (6.2\% on average) and take more steps (1.8\% on average). {Claude Haiku 4.5} is the only exception, showing marginal decreases in both. These results suggest that agents compensate for missing information through additional exploration, but that this extra effort is insufficient to recover the lost resolution rate. For most models, realistic requests are therefore both less successful and more expensive.

More importantly, model rankings are not preserved (Figure~\ref{fig:swebench_vs_realswe}). {MiMo V2.5 Pro} moves from fourth place on the original inputs to second place under \sysb{}, overtaking {Qwen3.7 Plus} and {DeepSeek V4 Flash}; the shift in the MiMo--Qwen gap between the two conditions is significant ($+3.7$ pp, 95\% CI $[+0.7, +7.3]$). {MiMo V2.5 Pro} is also roughly 2.5$\times$ cheaper per task than {Qwen3.7 Plus} (6.5 versus 16.1 cents). This suggests that the original leaderboard, taken at face value, would have steered users toward a model that is more expensive without being measurably better under realistic requests.

Realistic inputs also reshape the performance distribution in two ways. Among the four models clustered near the top of the original leaderboard ({Qwen3.7 Plus}, {DeepSeek V4 Flash}, {MiMo V2.5 Pro}, and {MiMo V2.5}), the performance range nearly doubles from 1.7 to 3.3 pp. At the same time, because the strongest models lose more in absolute terms, the overall strongest-to-weakest range narrows from 19.8 to 15.8 pp. Realistic inputs thus alter model separation rather than scaling all scores uniformly: models that looked interchangeable pull apart, while the field as a whole becomes more compressed.

Finally, the benchmark--reality gap varies substantially by task type. Bug-fix resolution decreases by 9.1 pp on average, whereas feature-request resolution decreases by only 3.7 pp. This suggests that underspecified user requests are especially challenging for bug repair, for which a precise description of the intended behavior may be critical. We test this explanation directly through the field-level analysis below.

\begin{table}[t]
\caption{Resolution rates and agent steps across information-field configurations for four models. $\Delta_{\mathrm{orig}}$ denotes the change from the original input, and $\Delta_{\mathrm{prev}}$ the change from the preceding configuration.}
\label{tab:field-ablation}
\centering
\fontsize{6.8}{8}\selectfont
\setlength{\tabcolsep}{1pt}
\begin{tabular*}{\textwidth}{@{} ll @{\extracolsep{\fill}} *{16}{c} @{}}
\toprule
& & \multicolumn{4}{c}{\textbf{DeepSeek V4 Pro}} & \multicolumn{4}{c}{\textbf{DeepSeek V4 Flash}} & \multicolumn{4}{c}{\textbf{MiMo V2.5 Pro}} & \multicolumn{4}{c}{\textbf{MiMo V2.5}} \\
\cmidrule(lr){3-6} \cmidrule(lr){7-10} \cmidrule(lr){11-14} \cmidrule(lr){15-18}
& \multicolumn{1}{c}{Fields} & Rate & $\Delta_{\mathrm{orig}}$ & $\Delta_{\mathrm{prev}}$ & Steps & Rate & $\Delta_{\mathrm{orig}}$ & $\Delta_{\mathrm{prev}}$ & Steps & Rate & $\Delta_{\mathrm{orig}}$ & $\Delta_{\mathrm{prev}}$ & Steps & Rate & $\Delta_{\mathrm{orig}}$ & $\Delta_{\mathrm{prev}}$ & Steps \\
\midrule
\multirow{7}{*}{\shortstack[c]{Bug fix\\($n$=192)}} & Orig. & 59.4$_{\pm1.4}$ & -- & -- & 42.2 & 56.2$_{\pm1.0}$ & -- & -- & 46.1 & 54.9$_{\pm1.6}$ & -- & -- & 45.4 & 54.5$_{\pm1.3}$ & -- & -- & 45.4 \\
 & \texttt{PDREA} & 57.8$_{\pm3.3}$ & $-$1.6 & $-$1.6 & 43.1 & 56.4$_{\pm0.3}$ & $+$0.2 & $+$0.2 & 45.7 & 55.7$_{\pm0.9}$ & $+$0.9 & $+$0.9 & 45.1 & 55.0$_{\pm0.6}$ & $+$0.5 & $+$0.5 & 44.5 \\
 & \texttt{PDRE} & 55.9$_{\pm1.1}$ & $-$3.5 & $-$1.9 & 42.2 & 57.1$_{\pm1.3}$ & $+$0.9 & $+$0.7 & 45.8 & 53.3$_{\pm1.3}$ & $-$1.6 & $-$2.4 & 44.9 & 51.6$_{\pm0.5}$ & $-$3.0 & $-$3.5 & 45.0 \\
 & \texttt{PDR} & 56.6$_{\pm1.5}$ & $-$2.8 & $+$0.7 & 42.0 & 54.9$_{\pm0.6}$ & $-$1.4 & $-$2.3 & 46.9 & 54.2$_{\pm1.4}$ & $-$0.7 & $+$0.9 & 45.0 & 54.9$_{\pm0.8}$ & $+$0.3 & $+$3.3 & 44.4 \\
 & \texttt{PD} & 56.6$_{\pm2.0}$ & $-$2.8 & $\hphantom{-}$0.0 & 42.7 & 52.8$_{\pm1.6}$ & $-$3.5 & $-$2.1 & 47.7 & 55.4$_{\pm1.7}$ & $+$0.5 & $+$1.2 & 46.3 & 52.8$_{\pm1.1}$ & $-$1.7 & $-$2.1 & 45.5 \\
 & \texttt{P} & 48.1$_{\pm1.3}$ & \textbf{\boldmath$-$11.3} & \textbf{\boldmath$-$8.5} & 44.7 & 45.3$_{\pm0.9}$ & \textbf{\boldmath$-$10.9} & \textbf{\boldmath$-$7.5} & 48.1 & 46.5$_{\pm1.8}$ & \textbf{\boldmath$-$8.3} & \textbf{\boldmath$-$8.8} & 45.9 & 45.7$_{\pm1.7}$ & \textbf{\boldmath$-$8.9} & \textbf{\boldmath$-$7.2} & 45.7 \\
 & \texttt{PA} & 50.5$_{\pm2.4}$ & $-$8.9 & $+$2.4 & 42.8 & 47.4$_{\pm0.9}$ & $-$8.9 & $+$2.1 & 46.9 & 51.6$_{\pm0.9}$ & $-$3.3 & $+$5.0 & 45.5 & 47.9$_{\pm1.0}$ & $-$6.6 & $+$2.3 & 44.5 \\
\midrule
\multirow{5}{*}{\shortstack[c]{Feature\\request\\($n$=189)}} & Orig. & 48.3$_{\pm0.3}$ & -- & -- & 43.0 & 43.0$_{\pm2.5}$ & -- & -- & 49.4 & 43.2$_{\pm2.1}$ & -- & -- & 50.5 & 42.2$_{\pm1.6}$ & -- & -- & 49.4 \\
 & \texttt{PMA} & 45.5$_{\pm1.8}$ & $-$2.8 & \textbf{\boldmath$-$2.8} & 43.9 & 43.7$_{\pm2.0}$ & $+$0.7 & $+$0.7 & 50.3 & 42.0$_{\pm0.3}$ & $-$1.2 & $-$1.2 & 52.6 & 38.3$_{\pm1.3}$ & $-$3.9 & \textbf{\boldmath$-$3.9} & 49.4 \\
 & \texttt{PM} & 45.7$_{\pm1.3}$ & $-$2.6 & $+$0.2 & 44.9 & 44.1$_{\pm1.6}$ & $+$1.1 & $+$0.4 & 49.5 & 42.2$_{\pm2.9}$ & $-$1.1 & $+$0.2 & 51.4 & 40.2$_{\pm1.9}$ & $-$1.9 & $+$1.9 & 48.7 \\
 & \texttt{P} & 42.9$_{\pm0.9}$ & \textbf{\boldmath$-$5.5} & \textbf{\boldmath$-$2.8} & 44.7 & 37.0$_{\pm0.5}$ & \textbf{\boldmath$-$6.0} & \textbf{\boldmath$-$7.1} & 51.3 & 40.2$_{\pm1.4}$ & \textbf{\boldmath$-$3.0} & \textbf{\boldmath$-$1.9} & 52.1 & 38.4$_{\pm0.8}$ & $-$3.7 & $-$1.8 & 49.8 \\
 & \texttt{PA} & 43.6$_{\pm0.8}$ & $-$4.8 & $+$0.7 & 44.9 & 41.1$_{\pm2.0}$ & $-$1.9 & $+$4.1 & 50.1 & 40.7$_{\pm1.9}$ & $-$2.5 & $+$0.5 & 52.4 & 36.7$_{\pm0.6}$ & \textbf{\boldmath$-$5.5} & $-$1.8 & 50.5 \\
\bottomrule
\end{tabular*}
\end{table}

\subsection{Controlled Analysis of \sys{} Results}
\label{sec:controlled}

To further investigate why realistic requests reduce coding-agent performance, we conduct a controlled analysis using the variants provided by \sysf{} (Table~\ref{tab:field-ablation}). This analysis uses four models: {DeepSeek V4 Pro}, {DeepSeek V4 Flash}, {MiMo V2.5 Pro}, and {MiMo V2.5}, representing two model families at different scales.

\paragraph{Ablation design.}
\label{sec:ablation-design}
Since the number of possible field combinations grows combinatorially, we organize the main analysis around cumulative ablation levels that begin with the most complete request and remove one field at a time. For bug fixes, the sequence is \texttt{[PDREA]} $\rightarrow$ \texttt{[PDRE]} $\rightarrow$ \texttt{[PDR]} $\rightarrow$ \texttt{[PD]} $\rightarrow$ \texttt{[P]}; for feature requests, it is \texttt{[PMA]} $\rightarrow$ \texttt{[PM]} $\rightarrow$ \texttt{[P]}. Each transition removes the rightmost field, allowing us to measure the effect of progressively reducing the available information. The all-field variants, \texttt{[PDREA]} and \texttt{[PMA]}, preserve the information contained in the original problem statement and differ only in linguistic style. Finally, because \texttt{[PA]} accounts for a large share of real requests in \textsc{SWE-chat}, we include it as an additional comparison condition. To verify that the observed effects are not specific to this cumulative ablation path, we evaluate a broader set of field combinations in Appendix~\ref{app:context-robustness}, focusing in particular on whether the effects of \texttt{[D]} and \texttt{[M]} persist when combined with other fields.

\paragraph{Effect of linguistic style.}
\label{sec:style}
Comparing the original problem statements with the all-field rephrased variants (i.e., \texttt{[PDREA]} and \texttt{[PMA]}) isolates the effect of linguistic style while holding information constant. Linguistic style has only a small, model-dependent effect: for bug fixes, the change ranges from $-1.6$ to $+0.9$ pp and averages 0.0 pp; for feature requests, it ranges from $-3.9$ to $+0.7$ pp and averages $-1.8$ pp. None of the eight style contrasts is significant (Holm-adjusted $p{\ge}.35$). Some models even improve after rephrasing, and no consistent direction emerges across models or task types. Overall, linguistic rephrasing alone does not systematically affect coding-agent performance; rather, as shown below, \emph{what} a request contains matters substantially more than \emph{how} it is expressed.

\paragraph{Information-field ablation.}
\label{sec:results}

For bug fixes, progressively removing \texttt{[A]}, \texttt{[E]}, and \texttt{[R]} changes resolution rate by only 1.8 pp on average. No removal step yields a significant decrease (Holm-adjusted $p{\ge}.44$), and the only significant change is an \emph{improvement} ($+$6.2 pp for {MiMo V2.5}, $p{=}.017$; Table~\ref{tab:main-statistical-analysis}), indicating that there is no noticeable or consistent effect of these information fields.

On the other hand, Desired Behavior \texttt{[D]} is markedly different. Removing \texttt{[D]} from \texttt{[PD]} lowers the resolution rate by 7.1--8.9 pp (8.0 pp on average; significant for all four models, Holm-adjusted $p{<}.01$). This loss is more than four times the combined decrease from removing \texttt{[A]}, \texttt{[E]}, and \texttt{[R]}. Although reproduction code and environment details may be essential for particular bugs, their average contribution along this ablation path is small, whereas Desired Behavior consistently provides substantial value.

Feature requests show similar concentration around Motivation \texttt{[M]}. Comparing \texttt{[PM]} with \texttt{[P]}, removing Motivation reduces resolution for all four models by 3.4 pp on average, and \texttt{[M]} is the only feature-request field whose average effect is distinguishable from zero (95\% CI $[+0.2, +8.2]$ pp). The effect is smaller and more model-dependent than that of Desired Behavior for bug fixes: DeepSeek V4 Flash loses 7.1 pp (Holm-adjusted $p{=}.001$), whereas the other models lose 1.8--2.8 pp (n.s.). Motivation is therefore valuable on average, but not as uniformly decisive as Desired Behavior. We attribute this pattern to the nature of the task type. In a feature request, the problem statement itself already describes the desired behavior, so the benefit that \texttt{[D]} provides for bug fixes comes built into \texttt{[P]}, and Motivation is the main information a user can still add, with a correspondingly smaller marginal value. Additional combinations reported in Appendix~\ref{app:context-robustness} confirm that the benefits of \texttt{[D]} and \texttt{[M]} persist across different surrounding information fields.

\paragraph{What users provide vs.\ what agents need.}
As established in \S\ref{sec:mismatch}, real users overwhelmingly submit \texttt{[P]} or \texttt{[PA]} requests (Figure~\ref{fig:Overall}), often omitting high-value fields such as Desired Behavior and Motivation while providing lower-value residual context. In fact, \texttt{[PD]} outperforms \texttt{[PA]} across all four models by 3.8--6.1 pp on bug-fix tasks, while \texttt{[PM]} outperforms \texttt{[PA]} by 1.4--3.5 pp on feature-request tasks. These results show that prompt length alone cannot precisely characterize the benchmark--reality gap: performance depends more on \textit{which} information a request contains than on \textit{how much} information it contains.
\section{Discussion and Limitations}

\paragraph{Actionable guidance for information-sparse prompts.}
The practical implication of our analysis is not that users should write benchmark-style issue reports for every software engineering request. Rather, users should specify the Desired Behavior when requesting a bug fix and the Motivation when requesting a new feature. In practice, however, users may omit this information because they do not recognize its value or because the intended behavior or motivation is itself unclear. Coding-agent interfaces can address this gap by explicitly asking targeted clarification questions before implementation. Alternatively, an agent may infer the missing Desired Behavior or Motivation from the request and repository context, incorporate it into the task specification, and then begin implementation. Such mechanisms could help coding agents better handle information-sparse requests and improve the performance users experience.

\paragraph{Model coverage.}
Although DeepSeek V4 Pro and MiMo V2.5 Pro are trillion-parameter-scale models, our evaluation does not include frontier systems that currently achieve leading performance on software engineering tasks, such as GPT-5.6~\citep{openai2026gpt56}, Opus 5~\citep{anthropic2026opus5}, and Kimi K3~\citep{moonshot2026kimik3}. Our findings may therefore not fully generalize to the strongest available coding models.

\paragraph{Multi-turn evaluation.}
Following \textsc{SWE-bench}, our evaluation is single-turn: the agent receives one task specification and completes the task without further interaction. It therefore does not capture how iterative clarification might mitigate sparse or ambiguous requests---a constraint of \textsc{SWE-bench}-style evaluation in general. Nevertheless, \sys{} enables a more structured analysis around which information should be prioritized during such interactions. Extending this to controlled conversational settings is important future work.
\section{Conclusion}
We introduced \sys{}, a benchmark and configurable framework for evaluating coding agents under realistic user requests. By characterizing the gap between real prompts and conventional \textsc{SWE-bench} problems, \sys{} enables controlled analysis of how information and linguistic style affect coding performance. Our results show that realistic inputs can substantially change measured performance, and that Desired Behavior and Motivation are particularly valuable signals. We hope \sys{} supports more realistic evaluation and better-informed coding-agent interfaces.

% ============================================================
% References
% ============================================================
\bibliographystyle{plainnat}
{\small
\bibliography{references}
}

% ============================================================
% Appendix
% ============================================================
\appendix
\lstdefinestyle{prompt}{%
  numbers=none,
  frame=single,
  framerule=0.5pt,
  framesep=6pt,
  xleftmargin=6pt,
  xrightmargin=6pt,
  breaklines=true,
  basicstyle={\fontsize{7}{8.5}\selectfont\ttfamily},
  aboveskip=8pt,
  belowskip=8pt}

\section{\sysb{} Details and Examples}
\label{app:composition}

\begin{figure}[ht]
\centering
\includegraphics[width=\textwidth]{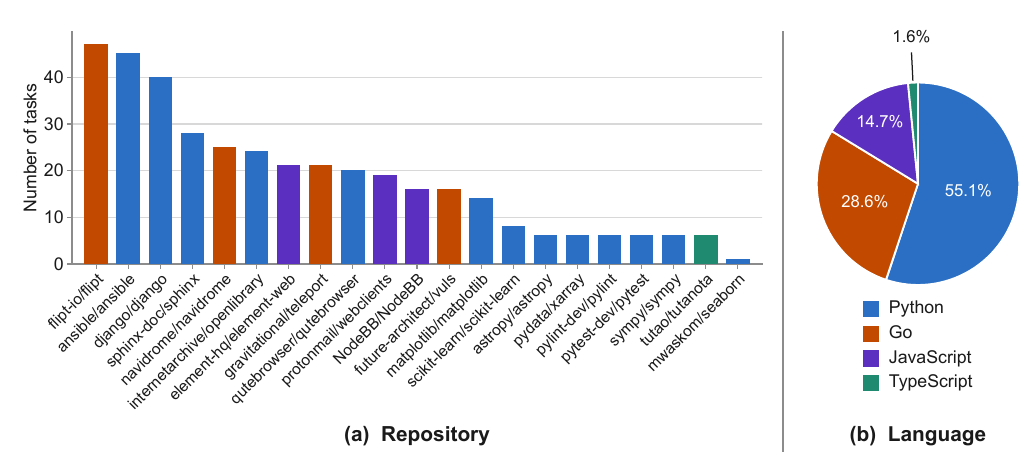}
\caption{Coverage of \sysb{} across repositories and programming languages.}
\label{fig:app-coverage}
\end{figure}

This section describes what \sysb{} contains: the benchmarks it is derived from, the repositories and programming languages it spans, the size of the gold patches its tasks require, a comparison of the selected tasks with the excluded ones, and examples of both the construction pipeline and the resulting tasks.

\subsection{Source Benchmarks}
\label{app:composition:benchmarks}

\sysb{} is derived from \textsc{SWE-bench Verified}~\citep{openai2024swebenchverified} and \textsc{SWE-bench Pro}~\citep{deng2025swebenchproaiagents}. We choose these two benchmarks because they represent widely used repository-level coding-agent evaluation.

\paragraph{\textsc{SWE-bench Verified}.}
\textsc{SWE-bench Verified} is a human-validated benchmark for standardized model comparison. Ninety-three professional software developers experienced in Python reviewed a random sample of 1,699 tasks from the \textsc{SWE-bench} test set~\citep{jimenez2024swebench} and removed those whose issue description is not well-specified or whose unit tests are inappropriately scoped, leaving 500 tasks. It contributes all 500, which come entirely from Python repositories.

\paragraph{\textsc{SWE-bench Pro}.}
\textsc{SWE-bench Pro} contributes 731 tasks that are more difficult and longer-horizon, drawn from larger and more complex repositories. Unlike \textsc{SWE-bench Verified}, it covers languages beyond Python (\S\ref{app:composition:coverage}). Every task carries \textit{Requirements} alongside the problem statement, and some also carry an \textit{Interface}; \S\ref{app:pipeline:decomposition} describes how we handle them.

\paragraph{Why not \textsc{DeepSWE}?}
We also considered \textsc{DeepSWE}~\citep{datacurve2026deepswe} as a source benchmark but could not use it. Each variant in \sys{} is built by redistributing the text of the original problem statement across the taxonomy fields without rewriting it (\S\ref{sec:constructing}), so a field the original does not contain cannot be produced. Supporting arbitrary combinations of information categories therefore requires every taxonomy field to be present in the source problem statement: the five fields \texttt{[P][D][R][E][A]} for bug fixes and the three fields \texttt{[P][M][A]} for feature requests (Table~\ref{tab:schema}). Applying this criterion to the 113 \textsc{DeepSWE} tasks leaves only 19 candidates, all of them feature requests. Moreover, {DeepSeek V4 Pro-preview}, the model with the highest resolution rate among the four models in Table~\ref{tab:field-ablation}, achieves a resolution rate of 0\% on these tasks even from the original problem statements, leaving no baseline against which to measure the effect of realistic requests.

\subsection{Repositories and Programming Languages}
\label{app:composition:coverage}

\sysb{} spans 21 repositories (Figure~\ref{fig:app-coverage}(a)) and four programming languages (Figure~\ref{fig:app-coverage}(b)). Python accounts for 55.1\% of the tasks. Because \textsc{SWE-bench Verified} contains only Python repositories, every task in the other three languages comes from \textsc{SWE-bench Pro}. Repository and language metadata follow the \textsc{SWE-bench Verified} and \textsc{SWE-bench Pro} releases on Hugging Face.

\subsection{Gold Patch Size}
\label{app:composition:size}

\begin{table}[t]
\caption{Size of the gold patch in \sysb{}.}
\label{tab:app-scale}
\centering
\small
\setlength{\tabcolsep}{8pt}
\begin{tabular}{@{}lccccc@{}}
\toprule
 & & \multicolumn{2}{c}{Files touched} & \multicolumn{2}{c}{Lines edited} \\
\cmidrule(l{2pt}r{2pt}){3-4} \cmidrule(l{2pt}r{2pt}){5-6}
 & Tasks & Mean & Median & Mean & Median \\
\midrule
\sysb{} & 381 & 4 & 3 & 134 & 62 \\
\quad $\hookrightarrow$ \textit{bug fixes} & 192 & 3 & 2 & 82 & 36 \\
\quad $\hookrightarrow$ \textit{feature requests} & 189 & 5 & 4 & 187 & 118 \\
\bottomrule
\end{tabular}
\end{table}

The size of the gold patch---the number of files it touches and the number of lines it adds or removes---indicates how much work a task requires. The gold patch for a task in \sysb{} touches 4 files and edits 134 lines on average (Table~\ref{tab:app-scale}).

Feature requests require larger patches than bug fixes: 5 files and 187 lines on average, compared with 3 files and 82 lines. This suggests that feature requests are the more difficult of the two task types, and indeed every model in Table~\ref{tab:main-results} has a lower resolution rate on feature requests than on bug fixes. As discussed in \S\ref{app:validation:selection}, this indicates that the tasks selected for \sys{} are not meaningfully easier and do not contain smaller gold patches compared to other tasks in \textsc{SWE-bench}, and that the selection is therefore not biased.

\subsection{Selection Bias}
\label{app:validation:selection}

Of the 1,229 tasks in the original pool---the 500 \textsc{SWE-bench Verified} tasks and the 729 \textsc{SWE-bench Pro} tasks that remain after removing the two whose task type is \textit{other} (Appendix~\ref{app:pipeline})---403 contain all required information categories, and excluding the 22 candidates that receive the lowest score on any critical criterion leaves 381. To check whether this selection favors easier or smaller tasks, or concentrates the benchmark on a few repositories, we compare the selected 381 tasks with the excluded 848 tasks on resolution rate, patch size, and repository distribution (Table~\ref{tab:app-selection-bias}). The resolution rate is that of DeepSeek V4 Pro on the original problem statements, reported as the mean over three runs with one standard deviation.

We measure repository distribution by the effective number of repositories,
\begin{equation*}
N_{\mathrm{eff}} = \frac{1}{\sum_{i=1}^{k} p_i^{2}},
\end{equation*}
where $p_i$ is the share of the group's tasks drawn from repository $i$ and $k$ is the number of repositories the group spans. The denominator is the Herfindahl--Hirschman index, so $N_{\mathrm{eff}}$ is the number of equally sized repositories that would give the same concentration; a larger value means the tasks are spread more evenly.

\begin{figure}[t]
\centering
\includegraphics[width=\textwidth]{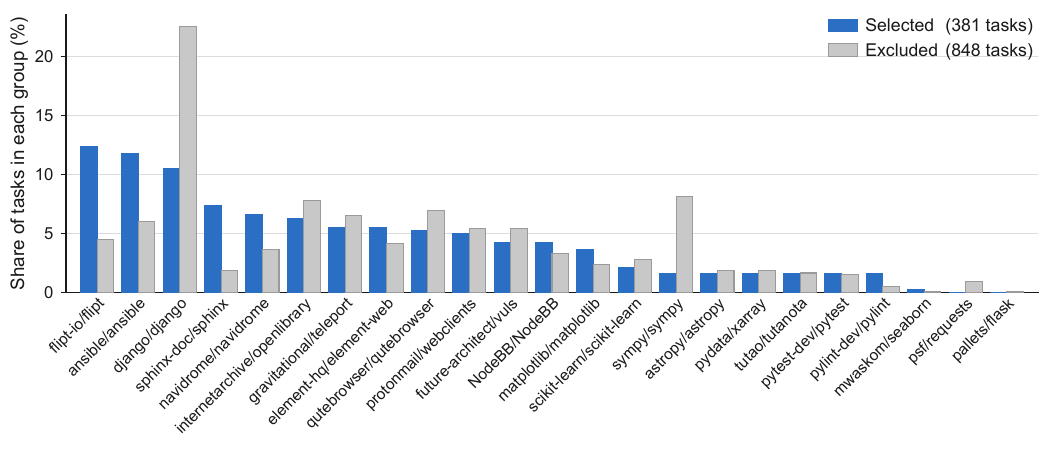}
\caption{Each repository's share of the selected tasks and of the excluded tasks.}
\label{fig:app-selection-repos}
\end{figure}

On none of the three measures are the selected tasks easier, smaller, or more concentrated in particular repositories. They are in fact harder (53.9\% against 63.0\%), larger (134 lines against 94), and spread more evenly (13.9 effective repositories against 11.0). Figure~\ref{fig:app-selection-repos} shows where that last difference comes from: \textit{django/django} alone accounts for 22.5\% of the excluded tasks, while no repository exceeds 12.3\% of the selected ones.

\begin{table}[t]
\caption{The selected 381 tasks compared with the excluded 848 tasks.}
\label{tab:app-selection-bias}
\centering
\small
\setlength{\tabcolsep}{10pt}
\begin{tabular}{@{}lcccccc@{}}
\toprule
 & & & \multicolumn{2}{c}{Lines edited} & \multicolumn{2}{c}{Repositories} \\
\cmidrule(l{2pt}r{2pt}){4-5} \cmidrule(l{2pt}r{2pt}){6-7}
 & Tasks & Resolution rate (\%) & Mean & Median & Count & Effective, $N_{\mathrm{eff}}$ \\
\midrule
Selected & 381 & $53.9_{\pm0.6}$ & 134 & 62 & 21 & 13.9 \\
Excluded & 848 & $63.0_{\pm1.2}$ & 94 & 38 & 23 & 11.0 \\
\bottomrule
\end{tabular}
\end{table}

\subsection{Example Tasks}
\label{app:composition:examples}

\begin{figure}[ht]
\centering
\includegraphics[width=\textwidth]{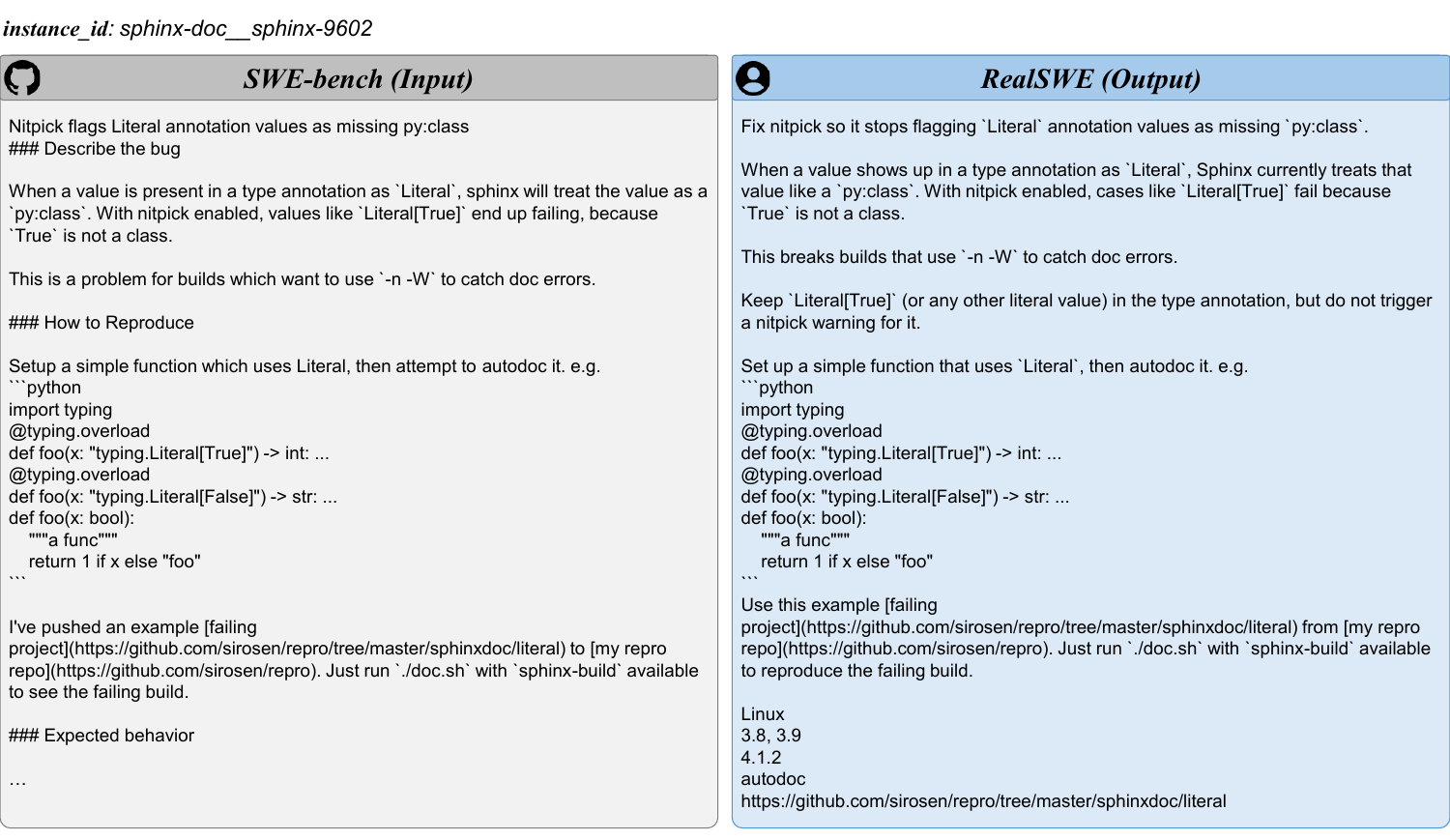}
\caption{The input and output of the construction pipeline for a single task. The original carries all five bug-fix fields of Table~\ref{tab:schema}, and the all-field variant \texttt{[PDREA]} preserves that information. It differs only in linguistic style---casual, imperative, confident, and non-first-person---as set in Appendix~\ref{app:pipeline:rephrasing}.}
\label{fig:app-before-after}
\end{figure}

\begin{figure}[t]
\centering
\includegraphics[width=\textwidth]{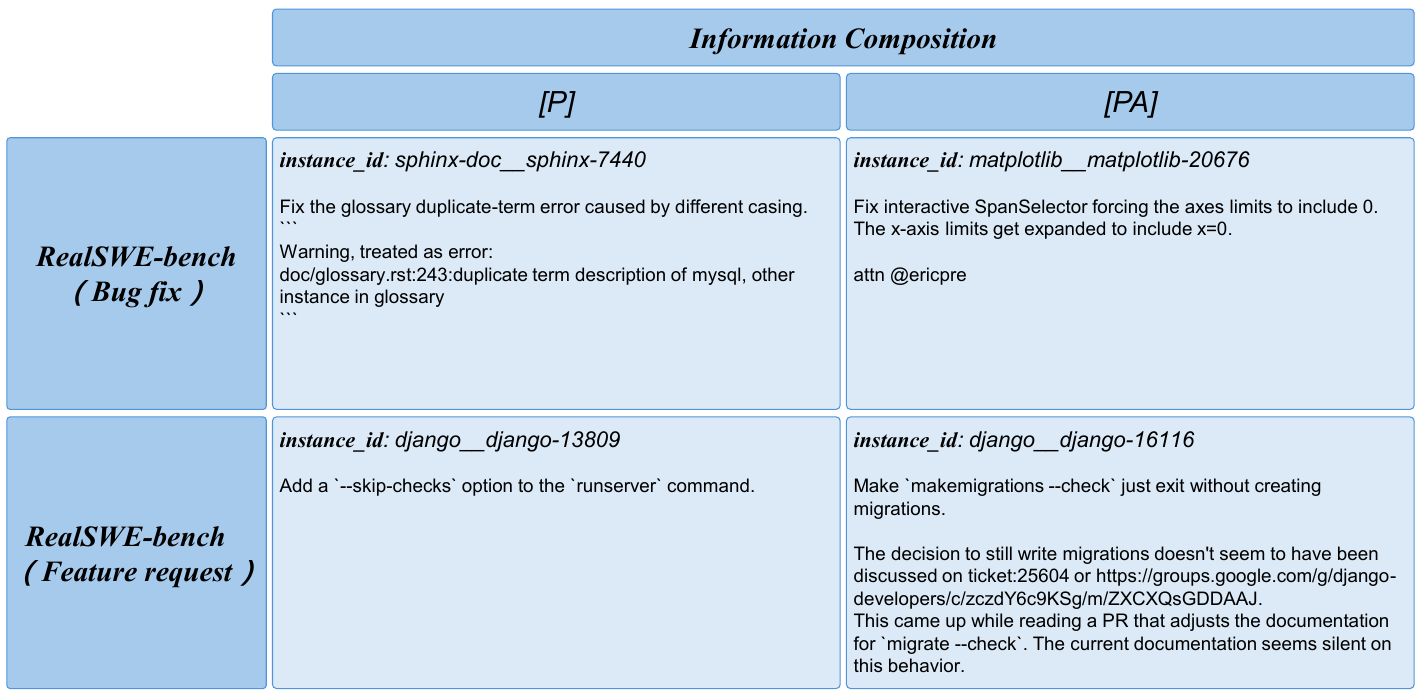}
\caption{Four tasks from \sysb{}, one for each combination of task type (bug fix, feature request) and information composition (\texttt{[P]}, \texttt{[PA]}).}
\label{fig:app-examples}
\end{figure}

Figure~\ref{fig:app-before-after} shows the input and output of the construction pipeline for a single task (\S\ref{sec:constructing}), and Figure~\ref{fig:app-examples} gives four example tasks from \sysb{}.

\FloatBarrier
\section{Construction Pipeline Details}
\label{app:pipeline}

This section reports what each step of the construction pipeline described in \S\ref{sec:constructing} produced, and why it is designed the way it is. Figure~\ref{fig:Our_Overview} shows the pipeline as a whole; Appendix~\ref{app:validation} validates each step and the resulting task quality.

\subsection{Task-Type Classification}
\label{app:pipeline:classification}

\begin{table}[b]
\caption{Distribution of task types in \textsc{SWE-bench Verified} and \textsc{SWE-bench Pro}.}
\label{tab:app-classification}
\centering
\small
\setlength{\tabcolsep}{10pt}
\begin{tabular}{@{}lccc@{}}
\toprule
 & \textsc{Verified} & \textsc{Pro} & Total \\
\midrule
\textit{Bug fix} & 432 (86.4\%) & 413 (56.5\%) & 845 (68.6\%) \\
\textit{Feature request} & \phantom{0}68 (13.6\%) & 316 (43.2\%) & 384 (31.2\%) \\
\textit{Other} & \phantom{00}0 \phantom{0}(0.0\%) & \phantom{00}2 \phantom{0}(0.3\%) & \phantom{00}2 \phantom{0}(0.2\%) \\
\midrule
Total & \phantom{0,}500 & \phantom{0,}731 & 1,231 \\
\bottomrule
\end{tabular}
\end{table}

We classify each task as a bug fix or a feature request using GPT-5.4~\citep{openai2026gpt54}. In \textsc{SWE-bench Pro}, a model is normally given the problem statement together with its \textit{Requirements} and \textit{Interface}; we classify from the problem statement alone, so that the classification reflects what the issue reports rather than the specification added for evaluation. Because some tasks are neither a bug fix nor a feature request, we offer \textit{other} as a third option.

Table~\ref{tab:app-classification} reports the counts. The two benchmarks differ substantially: 86.4\% of the \textsc{SWE-bench Verified} tasks are bug fixes, compared with 56.5\% for \textsc{SWE-bench Pro}.

Only two of the 1,231 tasks are assigned \textit{other}, both from \textsc{SWE-bench Pro}. Both are refactorings: they neither fix incorrect behavior nor add new or changed functionality. We exclude them and carry the remaining 1,229 tasks into the next step. That so few fall outside the two types indicates that bug fixes and feature requests cover essentially all tasks in both benchmarks.

\subsection{Information Decomposition}
\label{app:pipeline:decomposition}

This step divides each problem statement into the information fields of Table~\ref{tab:schema}. It redistributes the original text across those fields without rewriting it, so that decomposition does not alter the information the problem statement carries. Appendix~\ref{app:prompts} gives the prompt; Appendix~\ref{app:mismatch} reports the resulting field distributions in full.

Problem statements often retain template artifacts such as Markdown headings and HTML tags. These are not what actual users type, and leaving them in misdirects the model: when a user has written something that does not match what the template asked for, a section heading still in place leads the model to assign the line by the heading rather than by its content. We therefore remove these artifacts before decomposition, keeping everything the user wrote.

Unlike tasks in \textsc{SWE-bench Verified}, tasks in \textsc{SWE-bench Pro} are divided into three separate components: \textit{Problem Statement}, \textit{Requirements}, and \textit{Interface}. The \textit{Problem Statement} describes the problem to be solved; the \textit{Requirements} specify the expected behavior that the unit tests will check, and never include specific code implementation or leak the solution; and the \textit{Interface} provides code metadata, such as function signatures, that helps reduce false negatives during unit-test evaluation~\citep{deng2025swebenchproaiagents}.

To heuristically approximate a realistic coding environment, our pipeline merges the \textit{Problem Statement} and \textit{Requirements} into a single task description before processing it. The \textit{Interface} is excluded from the transformation pipeline and instead provided to the agent, when available, as an unmodified external document.

A task family can express arbitrary combinations of information categories only if every field of Table~\ref{tab:schema} is present---five for a bug fix and three for a feature request. The original problem statement supplies all of them in 302 problems; Appendix~\ref{app:mismatch} reports how often each field appears. Unlike the other fields, Environment Information \texttt{[E]} does not depend on what the issue author chose to write down: it is an objective property of the execution environment. For the 101 bug fixes whose statements carry only \texttt{[PDRA]} we therefore take it from the execution container released with its source benchmark, bringing the candidate set to 403. We add it to the original problem statement of those tasks as well, so the original carries the same information as the variants. Quality control then reduces the 403 to the 381 task families we release (Appendix~\ref{app:validation}).

\subsection{Rephrasing in Real-User Style}
\label{app:pipeline:rephrasing}

Every variant in \sysb{} and in Table~\ref{tab:field-ablation} shares one style setting. Following the \textsc{SWE-chat} distributions in Figure~\ref{fig:Overall}, we target the majority category of each linguistic dimension: casual for formality, imperative for sentence type, confident for certainty, and non-first-person for perspective. Appendix~\ref{app:prompts} gives the prompt.

We fix all four dimensions this way, but \sysf{} leaves them open: researchers specify a task type, an information composition, and a linguistic style, and the framework assembles the corresponding dataset on demand (Figure~\ref{fig:Our_Overview}), supporting custom configurations and controlled ablations.

\section{\textsc{SWE-chat}: Prompt Collection and Filtering}
\label{app:swechat}

\begin{figure}[b]
\centering
\includegraphics[width=0.7\textwidth]{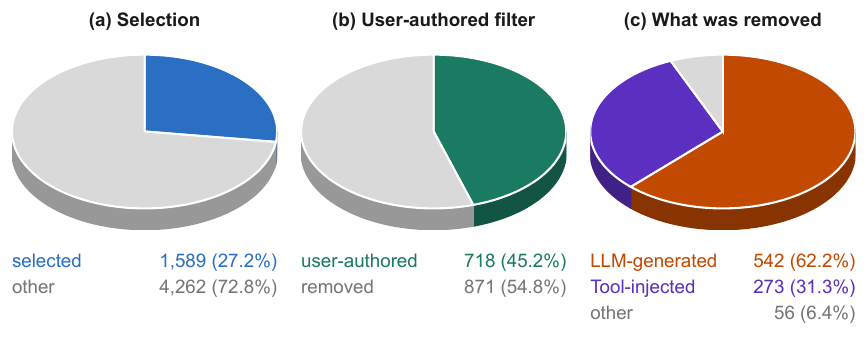}
\caption{Filtering \textsc{SWE-chat} down to the 718 prompts this paper analyzes.}
\label{fig:app-swechat-funnel}
\end{figure}

\textsc{SWE-chat}~\citep{baumann2026swechat} collects the session transcripts that the \textit{Entire} CLI records, with the developer's consent, for coding agents including Claude Code, OpenCode, Gemini CLI, Cursor, and Factory AI Droid. The snapshot we use contains 5,851 sessions (the \textsc{SWE-chat} paper reports more than 6,000). Because \textsc{SWE-bench} typically evaluates an agent from a single problem statement without further interaction, we retain only the first user request from each session. \textsc{SWE-chat} annotates each user prompt with one of eight intents; we keep the two that correspond to our task types, \textit{create new code} and \textit{debug}, which leaves 1,589 requests (Figure~\ref{fig:app-swechat-funnel}(a)).

Reviewing these requests, we found that a prompt recorded in a user turn is not necessarily written by the user. Text produced by an agent and wrapping injected by a tool sit in the user's place, and together they account for more than half of the set (Figure~\ref{fig:app-swechat-funnel}(b)). We are measuring how users actually write their requests, so we remove them.

Of the 871 removed requests, 542 (62.2\%) are LLM-generated and 273 (31.3\%) are tool-injected (Figure~\ref{fig:app-swechat-funnel}(c)). A single source dominates each: 523 of the 542 are plan-mode output from Claude Code, and 221 of the 273 are prompts wrapped by Conductor. The remaining 56 are personal templates whose source we could not verify and requests that give only an issue URL or a slash command, leaving the task itself outside the prompt.

The remaining 718 are the user-authored prompts this paper analyzes. We apply the information taxonomy and the linguistic dimensions to them to measure what real requests contain and how they are written; Appendix~\ref{app:mismatch} reports the resulting distributions.
\section{Benchmark--Reality Mismatch}
\label{app:mismatch}

\subsection{Information Composition}
\label{app:mismatch:composition}

\begin{figure}[t]
\centering
\includegraphics[width=0.8\textwidth]{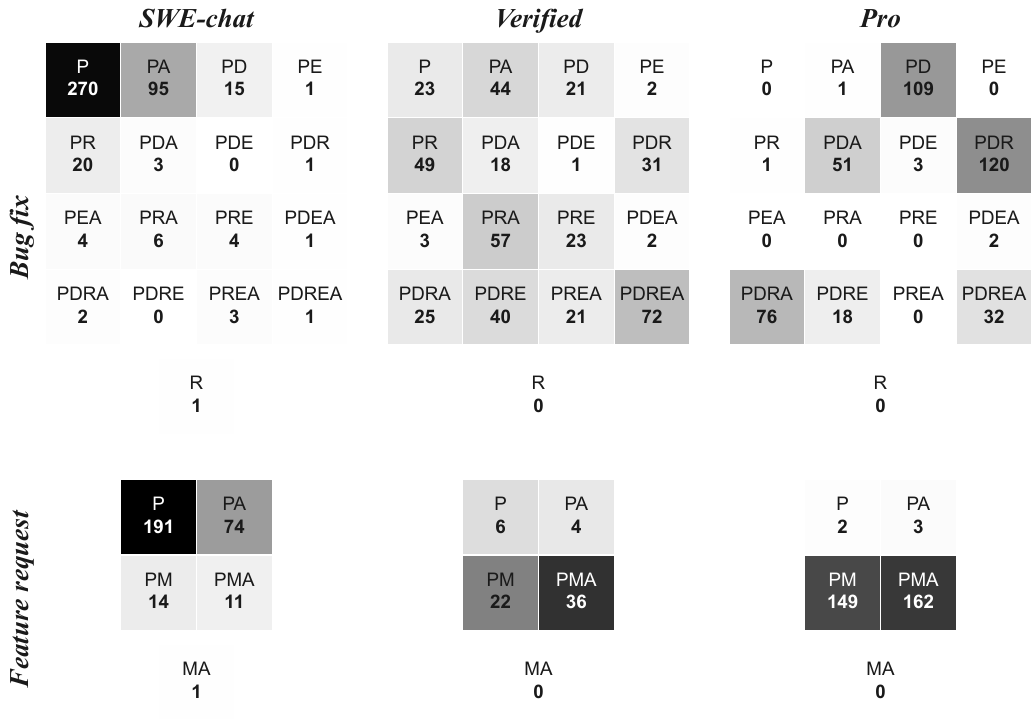}
\caption{Full distributions of information composition across \textsc{SWE-chat}, \textsc{SWE-bench Verified}, and \textsc{SWE-bench Pro}; each cell gives the number of prompts, and darker cells indicate larger shares.}
\label{fig:combo-appendix}
\end{figure}

Problems in \textsc{SWE-bench Verified} and \textsc{Pro} are information-rich (Figure~\ref{fig:combo-appendix}). Among the 845 bug fixes the most common combinations are \texttt{[PDR]} (17.9\%) and \texttt{[PD]} (15.4\%); the top two cover only a third, and the remainder spreads over fourteen further combinations. The 384 feature requests are almost entirely \texttt{[PMA]} (51.6\%) and \texttt{[PM]} (44.5\%), which together account for 96.1\%. Only 7\% of problems consist of \texttt{[P]} or \texttt{[PA]} alone (8.0\% for bug fixes, 3.9\% for feature requests).

Real requests run the other way. \texttt{[P]} and \texttt{[PA]} account for 88\% of the 718 prompts (85.5\% of bug-fix and 91.1\% of feature requests). A prompt carries 1.4 information fields on average, half the 2.9 of a benchmark problem, and 64.2\% consist of the problem statement \texttt{[P]} and nothing else; the same figure for the benchmark is 2.5\%.

Reading field by field shows where the gap sits (Table~\ref{tab:app-field-rates}). \texttt{[P]} is present in almost every prompt on both sides, so the difference lies entirely in the supporting fields. Reproduction Steps \texttt{[R]} appear in 8.9\% of real requests against 66.9\% of benchmark problems, and Environment Information \texttt{[E]} in 3.3\% against 25.9\%. The widest gaps belong to Desired Behavior \texttt{[D]} and Motivation \texttt{[M]}, the fields that raise the resolution rate most for bug fixes and for feature requests respectively (Table~\ref{tab:field-ablation}): 5.4\% against 73.5\% for \texttt{[D]}, and 8.9\% against 96.1\% for \texttt{[M]}. This gap likely leads benchmarks to overestimate coding performance in real-world settings.

\begin{table}[t]
\caption{Share of prompts (\%) containing each information field. Fields co-occur, so a column does not sum to 100.}
\label{tab:app-field-rates}
\centering
\small
\setlength{\tabcolsep}{6pt}
\begin{tabular}{@{}lcccc@{}}
\toprule
& \multicolumn{2}{c}{\textbf{Bug fix}} & \multicolumn{2}{c}{\textbf{Feature request}} \\
\cmidrule(l{2pt}r{2pt}){2-3} \cmidrule(l{2pt}r{2pt}){4-5}
& \textsc{SWE-chat} & Verified + Pro & \textsc{SWE-chat} & Verified + Pro \\
\midrule
Prompts & 427 & 845 & 291 & 384 \\
\midrule
\texttt{[P]} Problem Statement & 99.8 & 100.0 & 99.7 & 100.0 \\
\texttt{[D]} Desired Behavior & 5.4 & 73.5 & -- & -- \\
\texttt{[R]} Reproduction Steps & 8.9 & 66.9 & -- & -- \\
\texttt{[E]} Environment Information & 3.3 & 25.9 & -- & -- \\
\texttt{[M]} Motivation & -- & -- & 8.9 & 96.1 \\
\texttt{[A]} Additional Information & 26.9 & 47.8 & 29.6 & 53.4 \\
\bottomrule
\end{tabular}
\end{table}

\subsection{Linguistic Style}
\label{app:mismatch:style}

\begin{figure}[t]
\centering
\includegraphics[width=0.8\textwidth]{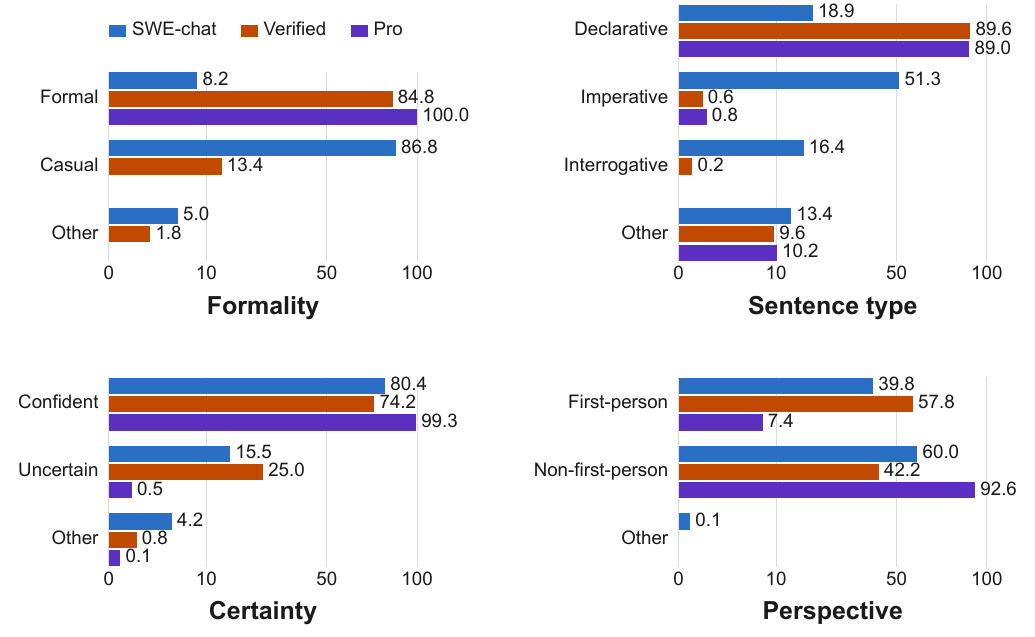}
\caption{Distributions of linguistic style along the four dimensions, as the share of prompts (\%). Blank cells are zero.}
\label{fig:style-grouped}
\end{figure}

The linguistic dimensions exhibit a less uniform pattern (Figure~\ref{fig:style-grouped}). Formality and sentence type divide the two sources most sharply: 86.8\% of real user requests are casual and 51.3\% are imperative, whereas 84.8\% of \textsc{SWE-bench Verified} and 100\% of \textsc{Pro} prompts are formal and approximately 89\% of prompts in both are declarative. \textsc{SWE-bench} prompts read as polished issue reports rather than conversational user requests. Certainty and perspective show no consistent separation.
\section{Additional Results and Analyses}
\label{app:results}

\subsection{Statistical Analysis}
\label{app:statistical-analysis}

Our statistical analysis examines whether the performance differences underlying the main comparisons are robust to variation across repeated runs and repositories, while quantifying their magnitude and uncertainty. We distinguish the roles of the reported statistics as follows.

\begin{center}
\small
\setlength{\tabcolsep}{8pt}
\renewcommand{\arraystretch}{1.15}
\begin{tabular}{@{}ll@{}}
\toprule
Statistic & What it shows \\
\midrule
Effect size $\Delta$ & Direction and magnitude \\
Repository-clustered 95\% CI & Effect-size uncertainty across repositories \\
Exact McNemar + Holm-adjusted $p$ & Model-level paired inference \\
Repository sign-flip $p$ & Aggregate and model-gap inference \\
\bottomrule
\end{tabular}
\end{center}

All analyses use matched comparisons in which different input conditions are applied to the same tasks. The main text reports the mean and sample standard deviation of three independent runs. Ranges and averages quoted in the main text are computed from unrounded values; table entries are rounded to one decimal place, so recomputing from the displayed entries may differ by 0.1 pp. For inference, we define a task as resolved under a condition when at least two of the three runs are resolved. For a given model, let $y_{icr}\in\{0,1\}$ denote the grader outcome for task $i$, condition $c$, and run $r$. The task-level outcome and paired effect are

\[
Y_{ic}=\mathbf{1}\!\left[\sum_{r=1}^{3}y_{icr}\ge 2\right],
\qquad
\widehat{\Delta}_{A\rightarrow B}
=\frac{100}{n}\sum_{i=1}^{n}\left(Y_{iB}-Y_{iA}\right).
\]

For model-level paired comparisons, we use a two-sided exact McNemar test and apply Holm correction when multiple comparisons jointly support one claim. Effects aggregated across models and the selected model-gap contrast instead use repository-level sign-flip tests. We obtain all 95\% confidence intervals from 10,000 repository-clustered bootstrap samples, preserving dependence among tasks from the same repository. The intervals are not multiplicity-adjusted and are not dual to either hypothesis test, so exclusion of zero need not coincide with $p<.05$.

\subsubsection{Statistical Support for the Main Findings}
\label{app:main-statistical-support}

Table~\ref{tab:main-statistical-analysis} summarizes the statistical evidence for the benchmark--reality gap, linguistic style, and information-field ablation.

\begin{table}[t]
\caption{Statistical support for the main comparisons. Positive $\Delta$ in the MiMo--Qwen row indicates a shift toward MiMo V2.5 Pro; the benchmark--reality and Motivation estimates aggregate seven and four models, respectively. The two linguistic-style rows form one eight-test Holm family, while each information-field comparison forms a separate four-model family. All effects use task-level majority outcomes, brackets report repository-clustered 95\% CIs, and bold denotes Holm-adjusted $p<.05$.}
\label{tab:main-statistical-analysis}
\centering
\footnotesize
\setlength{\tabcolsep}{3.5pt}
\renewcommand{\arraystretch}{1.15}

\begin{tabular*}{\textwidth}{@{}l@{\extracolsep{\fill}}lrrr@{}}
\toprule
\multicolumn{5}{l}{\textit{Panel A. Main-text comparisons}} \\
\midrule
Comparison & Scope & $\Delta$ (pp) & 95\% CI & \shortstack{Repository\\sign-flip $p$} \\
\midrule
\multirow{3}{*}{\shortstack[l]{Benchmark--reality gap\\Original $\rightarrow$ \sysb{}}}
& All tasks & $-6.6$ & $[-8.3,-5.0]$ & $<.001$ \\
& Bug fix & $-9.6$ & $[-13.1,-6.5]$ & $<.001$ \\
& Feature request & $-3.6$ & $[-7.6,-1.0]$ & $.008$ \\
\midrule
\shortstack[l]{MiMo V2.5 Pro margin over Qwen3.7 Plus\\Original $\rightarrow$ \sysb{}}
& All tasks & $+3.7$ & $[+0.7,+7.3]$ & $.040$ \\
\midrule
\shortstack[l]{Motivation effect\\\texttt{[P]} $\rightarrow$ \texttt{[PM]}}
& Feature request & $+3.8$ & $[+0.2,+8.2]$ & $.076$ \\
\bottomrule
\end{tabular*}

\vspace{15pt}

\begin{tabular*}{\textwidth}{@{}l@{\extracolsep{\fill}}cccc@{}}
\toprule
\multicolumn{5}{l}{\textit{Panel B. Linguistic style}} \\
\midrule
Comparison & \shortstack{DeepSeek\\V4 Pro} & \shortstack{DeepSeek\\V4 Flash} & \shortstack{MiMo\\V2.5 Pro} & \shortstack{MiMo\\V2.5} \\
\midrule
\shortstack[l]{Bug:\\Original $\rightarrow$ \texttt{[PDREA]}}
& \shortstack{$-1.6~[-4.3,+1.1]$\\Holm $p=1.000$}
& \shortstack{$-1.6~[-5.2,+2.5]$\\Holm $p=1.000$}
& \shortstack{$+1.0~[-2.3,+5.1]$\\Holm $p=1.000$}
& \shortstack{$0.0~[-3.0,+2.2]$\\Holm $p=1.000$} \\
\addlinespace[2.5pt]
\shortstack[l]{Feature:\\Original $\rightarrow$ \texttt{[PMA]}}
& \shortstack{$-2.1~[-8.8,+3.6]$\\Holm $p=1.000$}
& \shortstack{$+1.6~[-2.6,+5.4]$\\Holm $p=1.000$}
& \shortstack{$0.0~[-4.1,+4.9]$\\Holm $p=1.000$}
& \shortstack{$-5.8~[-10.7,-0.8]$\\Holm $p=.346$} \\
\midrule
\multicolumn{5}{l}{\textit{Panel C. Information-field ablation}} \\
\midrule
\shortstack[l]{Remove \texttt{[A]}:\\\texttt{[PDREA]} $\rightarrow$ \texttt{[PDRE]}}
& \shortstack{$-3.1~[-6.5,+0.5]$\\Holm $p=.438$}
& \shortstack{$+2.1~[-1.8,+6.0]$\\Holm $p=.848$}
& \shortstack{$-1.0~[-5.6,+2.9]$\\Holm $p=.848$}
& \shortstack{$-3.1~[-5.6,-0.6]$\\Holm $p=.438$} \\
\addlinespace[2.5pt]
\shortstack[l]{Remove \texttt{[E]}:\\\texttt{[PDRE]} $\rightarrow$ \texttt{[PDR]}}
& \shortstack{$+2.6~[-1.1,+6.2]$\\Holm $p=.801$}
& \shortstack{$-2.1~[-5.1,+0.9]$\\Holm $p=.801$}
& \shortstack{$-0.5~[-4.4,+2.9]$\\Holm $p=1.000$}
& \shortstack{$+6.2~[+1.4,+10.3]$\\Holm $p=\mathbf{.017}$} \\
\addlinespace[2.5pt]
\shortstack[l]{Remove \texttt{[R]}:\\\texttt{[PDR]} $\rightarrow$ \texttt{[PD]}}
& \shortstack{$-1.0~[-4.1,+1.9]$\\Holm $p=1.000$}
& \shortstack{$-1.0~[-5.6,+3.8]$\\Holm $p=1.000$}
& \shortstack{$+1.0~[-2.9,+4.9]$\\Holm $p=1.000$}
& \shortstack{$-3.6~[-7.8,+1.1]$\\Holm $p=.474$} \\
\addlinespace[2.5pt]
\shortstack[l]{Add \texttt{[D]}:\\\texttt{[P]} $\rightarrow$ \texttt{[PD]}}
& \shortstack{$+9.9~[+4.5,+16.0]$\\Holm $p=\mathbf{.002}$}
& \shortstack{$+8.9~[+3.8,+15.4]$\\Holm $p=\mathbf{.003}$}
& \shortstack{$+8.3~[+3.4,+12.8]$\\Holm $p=\mathbf{.005}$}
& \shortstack{$+6.8~[+2.2,+10.9]$\\Holm $p=\mathbf{.007}$} \\
\addlinespace[2.5pt]
\shortstack[l]{Add \texttt{[M]}:\\\texttt{[P]} $\rightarrow$ \texttt{[PM]}}
& \shortstack{$+0.5~[-4.4,+5.6]$\\Holm $p=1.000$}
& \shortstack{$+9.5~[+3.6,+16.8]$\\Holm $p=\mathbf{.001}$}
& \shortstack{$+3.2~[-1.2,+8.2]$\\Holm $p=.714$}
& \shortstack{$+2.1~[-2.0,+7.1]$\\Holm $p=1.000$} \\
\bottomrule
\end{tabular*}
\end{table}

The paired Original--\sysb{} comparisons show that the benchmark--reality gap persists when evaluated on the same tasks. Both task types exhibit a decrease, with a larger point estimate for bug fixes. The selected model contrast further supports the finding that realistic inputs can alter relative model comparisons, rather than merely lowering all scores uniformly.

Changing linguistic style while holding information fixed produces no consistent direction across models or task types. The information-field results instead distinguish the value of request content: removing \texttt{[A]}, \texttt{[E]}, or \texttt{[R]} produces no consistent degradation, whereas Desired Behavior \texttt{[D]} contributes robustly across all four models. Motivation \texttt{[M]} has a positive average contribution, but its evidence is more dependent on the model and repository. Taken together, the lack of a systematic style effect and the field-specific ablation pattern indicate that performance is more sensitive to \emph{which} information a request provides than to \emph{how} it is expressed or \emph{how much} information it provides.

\subsubsection{Robustness Across Information Contexts}
\label{app:context-robustness}

The cumulative ablation in the main text isolates each transition, but it does not by itself establish whether the key-field effects depend on that particular path. We therefore compare additional matched pairs that differ only in Desired Behavior or Motivation while holding the surrounding information fixed (Table~\ref{tab:context-robustness}).

\begin{table}[t]
\caption{Robustness of Desired Behavior \texttt{[D]} and Motivation \texttt{[M]} across surrounding information contexts on bug-fix and feature-request tasks, respectively. Additional context runs are limited to the two DeepSeek models due to computational cost. Descriptive effects are three-run mean resolution-rate differences; inferential effects use task-level majority outcomes with repository-clustered 95\% CIs, and all effects are in percentage points. For repeated main-path contrasts, inferential effect estimates and CIs are reused from Table~\ref{tab:main-statistical-analysis}, whereas Holm-adjusted $p$-values follow the 10-\texttt{[D]} and 4-\texttt{[M]} families defined here; bold denotes adjusted $p<.05$.}
\label{tab:context-robustness}
\centering
\footnotesize
\setlength{\tabcolsep}{4pt}
\renewcommand{\arraystretch}{1.15}

\begin{tabular*}{\textwidth}{@{}ll@{\extracolsep{\fill}}ccc@{}}
\toprule
\multirow{2}{*}{Matched pair} & \multirow{2}{*}{Model}
& Descriptive & \multicolumn{2}{c}{Inference} \\
\cmidrule(lr){3-3}\cmidrule(lr){4-5}
& & 3-run $\Delta$ & $\Delta$ [95\% CI] & Holm-adjusted $p$ \\
\midrule

\multicolumn{5}{l}{\textit{Desired Behavior \texttt{[D]}}} \\
\cmidrule(lr){1-5}

\multirow{2}{*}{\texttt{[P]} $\rightarrow$ \texttt{[PD]}}
& DeepSeek V4 Pro
& $+8.5$
& $+9.9~[+4.5,+16.0]$
& $\mathbf{.004}$ \\

& DeepSeek V4 Flash
& $+7.5$
& $+8.9~[+3.8,+15.4]$
& $\mathbf{.005}$ \\

\cmidrule(lr){1-5}

\multirow{2}{*}{\texttt{[PA]} $\rightarrow$ \texttt{[PDA]}}
& DeepSeek V4 Pro
& $+6.4$
& $+6.2~[+1.2,+11.5]$
& $.058$ \\

& DeepSeek V4 Flash
& $+7.6$
& $+8.3~[+3.5,+12.9]$
& $\mathbf{.012}$ \\

\cmidrule(lr){1-5}

\multirow{2}{*}{\texttt{[PR]} $\rightarrow$ \texttt{[PDR]}}
& DeepSeek V4 Pro
& $+8.7$
& $+9.4~[+5.6,+13.1]$
& $\mathbf{.002}$ \\

& DeepSeek V4 Flash
& $+7.3$
& $+6.2~[+1.2,+10.5]$
& $.051$ \\

\cmidrule(lr){1-5}

\multirow{2}{*}{\texttt{[PRE]} $\rightarrow$ \texttt{[PDRE]}}
& DeepSeek V4 Pro
& $+8.2$
& $+7.8~[+3.4,+12.1]$
& $\mathbf{.012}$ \\

& DeepSeek V4 Flash
& $+10.4$
& $+10.4~[+6.5,+14.6]$
& $\mathbf{.001}$ \\

\cmidrule(lr){1-5}

\multirow{2}{*}{\texttt{[PREA]} $\rightarrow$ \texttt{[PDREA]}}
& DeepSeek V4 Pro
& $+6.9$
& $+8.9~[+5.2,+12.8]$
& $\mathbf{.001}$ \\

& DeepSeek V4 Flash
& $+7.5$
& $+5.7~[+0.7,+10.2]$
& $.058$ \\

\midrule

\multicolumn{5}{l}{\textit{Motivation \texttt{[M]}}} \\
\cmidrule(lr){1-5}

\multirow{2}{*}{\texttt{[P]} $\rightarrow$ \texttt{[PM]}}
& DeepSeek V4 Pro
& $+2.8$
& $+0.5~[-4.4,+5.6]$
& $1.000$ \\

& DeepSeek V4 Flash
& $+7.1$
& $+9.5~[+3.6,+16.8]$
& $\mathbf{.001}$ \\

\cmidrule(lr){1-5}

\multirow{2}{*}{\texttt{[PA]} $\rightarrow$ \texttt{[PMA]}}
& DeepSeek V4 Pro
& $+1.9$
& $+3.7~[-2.2,+9.6]$
& $.430$ \\

& DeepSeek V4 Flash
& $+2.6$
& $+1.6~[-2.2,+6.5]$
& $1.000$ \\

\bottomrule
\end{tabular*}
\end{table}

Desired Behavior has a positive point estimate in every surrounding information context considered. Although the strength of individual comparisons varies after multiplicity correction, its benefit is not confined to the cumulative ablation path used in the main analysis.

Motivation likewise retains a positive direction with and without \texttt{[A]}, but with greater uncertainty and model variation. Overall, the direction of both key-field effects persists across surrounding information contexts, while their magnitude and precision remain context- and model-dependent.

\FloatBarrier 
\section{Validation and Quality Control}
\label{app:validation}

We validate two high-level claims from \S\ref{sec:validation}: \textit{(i)} the reliability of the construction pipeline and the quality of the resulting task families, and \textit{(ii)} the real-user linguistic style used to construct \sysb{}. Table~\ref{tab:human-validation-map} summarizes these validation blocks and the evidence used in each. Separately, \S\ref{app:validation:selection} audits whether the selection process disproportionately retains easier or smaller tasks or concentrates the selected pool in a few repositories.

\subsection{Pipeline Validation and Task Quality}
\label{app:construction-validation}

\begin{figure}[ht]
\centering
\includegraphics[width=\textwidth]{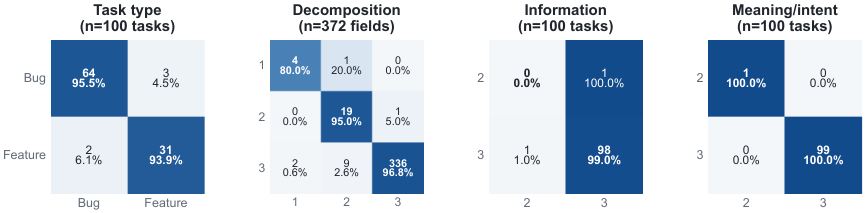}
\caption{Construction error structure. Rows are human labels and columns are pipeline or judge outputs; cells report counts and row-normalized percentages. Decomposition disagreements remain near the score boundary, and no human score-1 field is assigned judge score 3. Rephrasing disagreements occur only for information preservation between scores 2 and 3.}
\label{fig:construction-validation}
\end{figure}

We evaluate whether each construction stage makes only its intended transformation: \textit{i)} classification assigns each task to the correct task type, \textit{ii)} cleaning removes template artifacts without deleting user content, \textit{iii)} field assignment places each content unit in the appropriate taxonomy field, and \textit{iv)} rephrasing changes expression without altering task information or intent. We call a field assignment fully appropriate when every content unit matches its field definition. The complete scoring criteria are provided in Table~\ref{tab:validation-rubrics}. Agreement with human judgment (Table~\ref{tab:construction-validation}; Figure~\ref{fig:construction-validation}) supports applying the quality-control judge to all candidates; the resulting gate removes 22 critical decomposition failures and no rephrasing failures, leaving 381 task families. We additionally validate the \textsc{SWE-chat} field assignments that define the target information composition.

\begin{table}[ht]
\caption{Human-Validation map. Each row corresponds to a validation block in Sections~\ref{app:construction-validation} and~\ref{app:linguistic-style-validation}.}
\label{tab:human-validation-map}
\centering
{\small
\renewcommand{\arraystretch}{1.05}
\setlength{\tabcolsep}{6pt}
\begin{tabular}{@{}p{0.23\textwidth}p{0.42\textwidth}p{0.27\textwidth}@{}}
\toprule
{\raggedright\textbf{Validation block}\par}
& {\raggedright\textbf{Question}\par}
& {\raggedright\textbf{Evidence}\par} \\
\midrule
{\raggedright Pipeline validation and task quality (\ref{app:construction-validation})\par}
& {\raggedright Do the pipeline stages produce their intended outputs without critical failures?\par}
& {\raggedright Human audit + judge validation\par} \\
\addlinespace[6pt]
{\raggedright Linguistic-style validation (\ref{app:linguistic-style-validation})\par}
& {\raggedright Do the linguistic annotations reliably characterize real-user style, and does \sysb{} realize it?\par}
& {\raggedright Pairwise comparison + human re-annotation\par} \\
\bottomrule
\end{tabular}
}
\end{table}

\begin{table}[ht]
\caption{Operational three-point rubrics. Score 1 denotes a critical failure. The full-pool gate excludes a candidate if any non-empty decomposition field or rephrasing criterion receives score 1. Human annotators and the quality-control judge use the same definitions for these two stages.}
\label{tab:validation-rubrics}
\centering
{\small
\renewcommand{\arraystretch}{1.10}
\setlength{\tabcolsep}{4pt}
\begin{tabular}{@{}p{0.19\textwidth}p{0.235\textwidth}p{0.235\textwidth}p{0.235\textwidth}@{}}
\toprule
{\raggedright\textbf{Criterion}\par}
& {\centering\textbf{Score 3}\par}
& {\centering\textbf{Score 2}\par}
& {\centering\textbf{Score 1}\par} \\
\midrule
{\raggedright Cleaning: over-deletion\par}
& {\raggedright No user-authored content is removed.\par}
& {\raggedright Minor user-authored content is removed without affecting the task.\par}
& {\raggedright Task-relevant user content is removed.\par} \\
\addlinespace[4pt]
{\raggedright Cleaning: under-deletion\par}
& {\raggedright All target template artifacts are removed.\par}
& {\raggedright Minor template artifacts remain without obscuring the request.\par}
& {\raggedright Substantial template artifacts remain, or cleaning is effectively absent.\par} \\
\addlinespace[4pt]
{\raggedright Information-field assignment\par}
& {\raggedright Every sentence and content unit matches the definition of its assigned field.\par}
& {\raggedright Some content would fit another field better.\par}
& {\raggedright All or nearly all content is assigned to an inappropriate field.\par} \\
\addlinespace[4pt]
{\raggedright Rephrasing: information preservation\par}
& {\raggedright All task information and technical literals, including code, errors, tracebacks, and paths, are preserved; no task-relevant claim is added.\par}
& {\raggedright A minor addition or omission does not change the required solution, and technical literals remain intact.\par}
& {\raggedright Material information is added or removed, or a technical literal is altered.\par} \\
\addlinespace[4pt]
{\raggedright Rephrasing: meaning and intent preservation\par}
& {\raggedright Scope, causal relations, requirement strength, and implementation objective are unchanged.\par}
& {\raggedright A minor semantic shift or ambiguity leaves the implementation objective unchanged.\par}
& {\raggedright Scope, causal relations, requirement strength, or implementation objective is materially changed.\par} \\
\bottomrule
\end{tabular}
}
\end{table}

\begin{table}[ht]
\caption{Pipeline validation and task-quality results. Agreement percentages compare the two human annotators; parenthetical values report Cohen's \(\kappa\) or quadratic-weighted \(\kappa\) (QWK). Judge--human results use human consensus as the reference.}
\label{tab:construction-validation}
\centering
{\small
\renewcommand{\arraystretch}{1.08}
\setlength{\tabcolsep}{6pt}
\begin{tabular}{@{}p{0.24\textwidth}cp{0.41\textwidth}@{}}
\toprule
{\raggedright\textbf{Stage}\par}
& \textbf{Annotator agreement}
& {\raggedright\textbf{Validation result}\par} \\
\midrule
\multicolumn{3}{@{}l}{\textbf{Data preprocessing}} \\
\addlinespace[2pt]
{\raggedright Task-type classification\par}
& 88.0\% ($\kappa=0.75$)
& {\raggedright \textbf{95\%} correctly classified\par} \\
{\raggedright Template cleaning\par}
& 95--96\% (QWK 0.81--0.85)
& {\raggedright Critical failures: \textbf{1\%} user-content loss; \textbf{2\%} retained template artifacts\par} \\
\addlinespace[4pt]
\multicolumn{3}{@{}l}{\textbf{Information-composition annotation}} \\
\addlinespace[2pt]
{\raggedright \textsc{SWE-chat} field assignment\par}
& 91.2\% (QWK 0.40)
& {\raggedright \textbf{93.2\%} fully appropriate\par} \\
\addlinespace[4pt]
\multicolumn{3}{@{}l}{\textbf{Task-family construction}} \\
\addlinespace[2pt]
{\raggedright Field decomposition\par}
& 92.7\% (QWK 0.62)
& {\raggedright \textbf{97\%} judge--human match (macro-$F_1$ 0.83)\par} \\
{\raggedright Linguistic rephrasing\par}
& 95\% information / 94\% intent
& {\raggedright \textbf{99\%} judge--human match (macro-$F_1$ 0.75)\par} \\
\bottomrule
\end{tabular}
}
\end{table}

\FloatBarrier

\subsection{Linguistic-Style Validation}
\label{app:linguistic-style-validation}

We validate both the linguistic-dimension annotations used to characterize real-user style and the rephrasing used to construct \sysb{}.

\subsubsection{Real-User-Style Rephrasing}
\label{app:target-style-validation}

We validate whether the rephrasing used to construct \sysb{} reflects the real-user linguistic style characterized in \S\ref{sec:mismatch}. Pairwise judgments (Table~\ref{tab:target-style-validation}) favor the rephrased requests in formality and sentence type, the two dimensions with the clearest \textsc{SWE-chat}--\textsc{SWE-bench} contrast. The weaker effects in certainty and perspective mirror their less consistent source-level differences, together supporting the intended real-user-style rephrasing.

\begin{table}[ht]
\caption{Blind pairwise comparison of the restructured and rephrased requests used to construct \sysb{}. For each dimension, the rubric selects the candidate closer to the target category---casual, imperative, confident, or non-first-person---or a tie. Decision columns report the most common judgment and its share. To mitigate position bias, full-pool judge results average the original and swapped candidate orders after mapping both judgments back to their source labels. Human decisions aggregate the two annotators' 200 source-normalized judgments per dimension on a matched sample of 100 pairs; judge--human agreement averages exact agreement across both annotators and both candidate orders on the same sample.}
\label{tab:target-style-validation}
\centering
{\small
\renewcommand{\arraystretch}{1.05}
\setlength{\tabcolsep}{4pt}      
\begin{tabular}{@{}lp{0.23\textwidth}p{0.23\textwidth}cc@{}}  
\toprule
\textbf{Dimension}
& {\raggedright\textbf{Judge decision}\par}
& {\raggedright\textbf{Human decision}\par}
& \shortstack{\textbf{Human agreement}\\\textbf{($\kappa$)}}
& \shortstack{\textbf{Judge--human}\\\textbf{agreement}} \\
\midrule
Formality & {\raggedright Rephrased closer (80.8\%)\par} & {\raggedright Rephrased closer (82.0\%)\par} & 0.64 & 84.8\% \\
Sentence type & {\raggedright Rephrased closer (97.4\%)\par} & {\raggedright Rephrased closer (97.0\%)\par} & 0.92 & 97.5\% \\
Certainty & {\raggedright Rephrased closer (43.7\%)\par} & {\raggedright Rephrased closer (61.5\%)\par} & 0.83 & 69.3\% \\
Perspective & {\raggedright Tie (79.7\%)\par} & {\raggedright Tie (69.0\%)\par} & 0.75 & 91.5\% \\
\bottomrule
\end{tabular}
}
\end{table}

\subsubsection{Linguistic-Dimension Annotation}
\label{app:realism-target-validation}

\begin{figure}[ht]
\centering
\includegraphics[width=\textwidth]{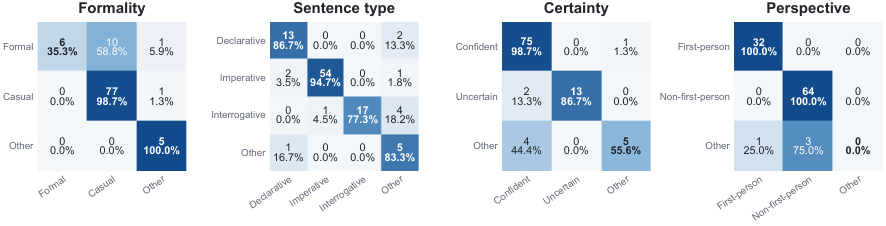}
\caption{\textsc{SWE-chat} annotation errors. Rows are human consensus and columns are classifier outputs; cells report counts and row-normalized percentages. Errors are concentrated in sparse minority labels, while the dominant casual and imperative classes are preserved.}
\label{fig:swechat-dimension-confusions}
\end{figure}

\begin{figure}[ht]
\centering
\includegraphics[width=\textwidth]{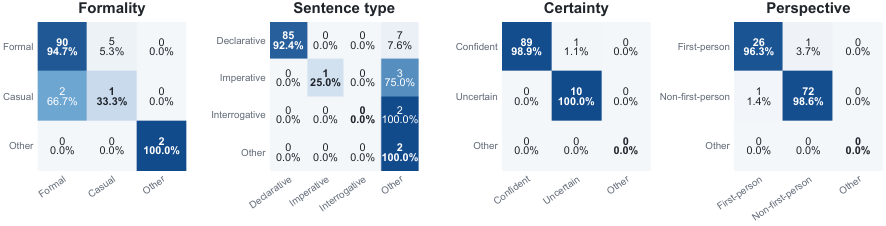}
\caption{\textsc{SWE-bench} annotation errors in the same layout and scale as Figure~\ref{fig:swechat-dimension-confusions}. The low sentence-type macro-F1 reflects minority-class errors rather than failure to identify the dominant declarative class.}
\label{fig:swebench-dimension-confusions}
\end{figure}

We validate whether the linguistic-dimension annotations underlying the benchmark--reality comparison reflect human judgment. Table~\ref{tab:realism-target-validation} shows close agreement with human annotations, while Figures~\ref{fig:swechat-dimension-confusions} and~\ref{fig:swebench-dimension-confusions} locate the remaining errors in sparse minority classes. Human re-annotation preserves the primary linguistic contrast used to define the real-user style: \textsc{SWE-chat} is predominantly casual and imperative, whereas \textsc{SWE-bench} is predominantly formal and declarative.

\begin{table}[ht]
\caption{Human validation of the linguistic-dimension annotations underlying the benchmark--reality comparison. IAA denotes raw inter-annotator agreement, with Cohen's \(\kappa\) in parentheses. Classifier accuracy and macro-\(F_1\) use human consensus as the reference.}
\label{tab:realism-target-validation}
\centering
{\small
\renewcommand{\arraystretch}{1.05}
\setlength{\tabcolsep}{8pt}
\begin{tabular}{@{}lcccc@{}}
\toprule
& \multicolumn{2}{c}{\textbf{\textsc{SWE-chat}}}
& \multicolumn{2}{c}{\textbf{\textsc{SWE-bench}}} \\
\cmidrule(lr){2-3}\cmidrule(lr){4-5}
\textbf{Dimension}
& \shortstack{\textbf{IAA}\\\textbf{($\kappa$)}}
& \shortstack{\textbf{Accuracy}\\\textbf{(macro-$F_1$)}}
& \shortstack{\textbf{IAA}\\\textbf{($\kappa$)}}
& \shortstack{\textbf{Accuracy}\\\textbf{(macro-$F_1$)}} \\
\midrule
Formality & 82\% (0.48) & 88\% (0.76) & 88\% (0.15) & 93\% (0.73) \\
Sentence type & 86\% (0.77) & 89\% (0.81) & 91\% (0.37) & 88\% (0.40) \\
Certainty & 92\% (0.79) & 93\% (0.85) & 90\% (0.61) & 99\% (0.97) \\
Perspective & 94\% (0.87) & 96\% (0.65) & 98\% (0.95) & 98\% (0.97) \\
\bottomrule
\end{tabular}
}
\end{table}

\FloatBarrier 
\section{Experimental Details}
\label{app:experimental-details}

Table~\ref{tab:models} reports the exact model snapshot, serving provider, and inference setting for every model used in this work. Models routed through OpenRouter carry the \texttt{openrouter/} prefix and are listed with their serving provider; the rest were accessed directly. The reasoning column gives the reasoning-effort setting, where \textit{default} denotes models that expose only an on/off toggle and were run with reasoning on; Claude Haiku 4.5 exposes a thinking-token budget instead of an effort level.

\begin{table}[ht]
\caption{Models used in this work.}
\label{tab:models}
\centering
\footnotesize
\setlength{\tabcolsep}{2pt}
\begin{tabular}{@{}llcll@{}}
\toprule
\textbf{Role} & \textbf{Model} & \textbf{Provider} & \textbf{Model ID} & \textbf{Reasoning} \\
\midrule
Pipeline & GPT-5.4 & -- & \texttt{gpt-5.4-2026-03-05} & low \\
\midrule
Evaluated models & DeepSeek V4 Pro-preview & -- & \texttt{deepseek/deepseek-v4-pro} & high \\
 & DeepSeek V4 Flash-preview & -- & \texttt{deepseek/deepseek-v4-flash} & high \\
 & MiMo V2.5 Pro & Xiaomi & \texttt{openrouter/xiaomi/mimo-v2.5-pro} & default \\
 & MiMo V2.5 & Xiaomi & \texttt{openrouter/xiaomi/mimo-v2.5} & default \\
 & Claude Haiku 4.5 & -- & \texttt{anthropic/claude-haiku-4-5-20251001} & budget 16{,}384 \\
 & Qwen3.7 Plus & Alibaba & \texttt{openrouter/qwen/qwen3.7-plus} & default \\
 & MiniMax M3 & MiniMax & \texttt{openrouter/minimax/minimax-m3} & default \\
\midrule
LLM-as-a-judge & GPT-5.6 Terra & -- & \texttt{gpt-5.6-terra} & low \\
\bottomrule
\end{tabular}
\end{table}

\FloatBarrier
\section{LLM Prompts}
\label{app:prompts}

This section provides the complete prompts used at every point where this paper relies on an LLM.

\subsection{Task-Type Classification}
\label{app:prompts:classification}

\begin{lstlisting}[style=prompt]
PROMPT_CLASSIFICATION = """
# Identity

You are an annotator. Your task is to classify a GitHub issue into one of three types: "bug", "feature", or "other".

# Task

You will receive one problem statement from a GitHub issue. Assign exactly one issue type:

- bug: A bug report. The issue describes a problem in existing functionality.
- feature: A feature request. The issue requests an addition or improvement to the codebase.
- other: The issue does not clearly fit either a bug report or a feature request.

# Rules

1. Your task is only to select the issue type. Do not modify, paraphrase, or extract any content from the problem statement.
2. Prefer bug report or feature request when the issue can reasonably be classified as either. Only assign other when neither a bug report nor a feature request applies.
"""
\end{lstlisting}

\subsection{Template Artifact Cleaning}
\label{app:prompts:cleaning}

\begin{lstlisting}[style=prompt]
PROMPT_CLEANING = """
# Identity

You are a text cleaner. Your task is to remove GitHub-template artifacts from a GitHub issue's problem statement while preserving all user-authored content verbatim.

# Task

You will receive a problem statement from a GitHub issue. Return it verbatim with all GitHub-template artifacts removed and all user-authored content preserved.
The cleaned text will be restructured into predefined fields by a downstream pipeline:

- Bug reports: describe_the_bug, expected_behavior, reproduction_code, environment, additional_context.
- Feature requests: desired_solution, problem_motivation, additional_context.

# Rules

1. The single criterion for removal is whether the content was authored by the user or was automatically inserted by the GitHub issue template.
2. Remove all template artifacts. These include any text that the issue template prompted or pre-filled, such as section labels naming a field the user was asked to complete, hidden instructions guiding the user, and boilerplate text referencing project conventions.
3. Preserve all user-authored content verbatim. This includes the user's own prose, code, error messages, references, and any markdown or headers the user added to organize their own writing.
4. When uncertain whether content is user-authored or a template artifact, preserve it.
5. If the problem statement contains no GitHub-template artifacts, return it unchanged.
"""
\end{lstlisting}

\subsection{Information Decomposition}
\label{app:prompts:decomposition}

\begin{lstlisting}[style=prompt]
_TEMPLATE_BUG = """
# Identity

You are an annotator. Your task is to classify each line of {owner} into five predefined fields, preserving every line verbatim.

# Task

You will receive {intro}. Classify each line of {noun} verbatim into these five fields:
- describe_the_bug: The core problem or the task to address, along with any output or error messages.
- expected_behavior: A clear and concise description of what should have happened instead of the bug.
- reproduction_code: Steps or code to reproduce the bug.
- environment: Runtime environment details, such as OS, language version, and package versions.
- additional_context: Any other context about the problem.

# Rules

1. Preserve every line of {noun} exactly. Do not alter spelling, punctuation, whitespace within a line, or code.
2. Every non-blank line of {noun} must appear in exactly one field. Do not duplicate a line across fields, and do not drop any line.
3. Lines within a single code block or traceback must be kept together in the same field. When multiple code blocks or tracebacks exist, each can be assigned to a different field based on which field best matches its content.
4. additional_context is the catch-all: content that does not clearly fit the other four fields goes here.
5. If a field has no content, use the JSON value null without quotation marks. Never use the strings "null", "None", "N/A", or any generated placeholder to represent an empty field.
"""

_TEMPLATE_FEATURE = """
# Identity

You are an annotator. Your task is to classify each line of {owner} into three predefined fields, preserving every line verbatim.

# Task

You will receive {intro}. Classify each line of {noun} verbatim into these three fields:
- desired_solution: A clear and concise description of the proposed change or desired behavior.
- problem_motivation: A clear and concise description of the current problem or limitation.
- additional_context: Any other context about the feature request.

# Rules

1. Preserve every line of {noun} exactly. Do not alter spelling, punctuation, whitespace within a line, or code.
2. Every non-blank line of {noun} must appear in exactly one field. Do not duplicate a line across fields, and do not drop any line.
3. Lines within a single code block or traceback must be kept together in the same field. When multiple code blocks or tracebacks exist, each can be assigned to a different field based on which field best matches its content.
4. additional_context is the catch-all: content that does not clearly fit the other two fields goes here.
5. If a field has no content, use the JSON value null without quotation marks. Never use the strings "null", "None", "N/A", or any generated placeholder to represent an empty field.
"""

_TEMPLATE = {"bug": _TEMPLATE_BUG, "feature": _TEMPLATE_FEATURE}

SUBJECTS = {
    "problem_statement": {
        "bug": {
            "owner": "a GitHub bug report's problem statement",
            "intro": "one problem statement from a GitHub bug report",
            "noun": "the problem statement",
        },
        "feature": {
            "owner": "a GitHub feature request's problem statement",
            "intro": "one problem statement from a GitHub feature request",
            "noun": "the problem statement",
        },
    },
    "requirements": {
        "bug": {
            "owner": "the bug-fix task's requirements",
            "intro": "the requirements of the bug-fix task",
            "noun": "the requirements",
        },
        "feature": {
            "owner": "the feature task's requirements",
            "intro": "the requirements of the feature task",
            "noun": "the requirements",
        },
    },
}
\end{lstlisting}

\subsection{Linguistic Rephrasing}
\label{app:prompts:rephrasing}

\begin{lstlisting}[style=prompt]
PROMPT_REPHRASING = """
# Identity
You are a rephraser. Your task is to rephrase each field of a GitHub issue into content that users actually send to an AI coding agent (such as Claude Code, Codex, or Cursor) in the real world.

# Input
You will receive a structured GitHub issue, which is either a bug report or a feature request.

Bug reports are structured with five fields:
- describe_the_bug
- expected_behavior
- reproduction_code
- environment
- additional_context

Feature requests are structured with three fields:
- desired_solution
- problem_motivation
- additional_context

# Task
Rephrase each field following the instructions below while preserving the original meaning and intent.

# Rephrasing Instructions

1. Formality
   If the vocabulary or expressions are not casual, convert them to casual ones.

2. Sentence type
   Prefer imperative types. Declarative and interrogative types are acceptable when more natural.

3. Certainty
   If the vocabulary or expressions are not confident, convert them to confident ones.

4. Perspective
   Prefer non-first-person perspectives. First-person perspectives are acceptable when more natural.

# Rules

1. Preserve code blocks, error messages, tracebacks, version numbers, file paths, and other technical content exactly as-is.

2. Preserve the original meaning and intent.
"""
\end{lstlisting}

\subsection{Decomposition Quality Judgment}
\label{app:prompts:judge-decomposition}

\begin{lstlisting}[style=prompt]
PROMPT_JUDGE_DECOMPOSITION = """
# Role
You are an evaluator assessing whether the content assigned to each field of a restructured GitHub issue matches that field's definition.

# Task
You will receive `restructured_ps`, a JSON object containing the fields of one restructured problem statement.

For each field, judge only whether the content placed in that field belongs there according to its definition.

## Field Definitions
{field_descriptions}

# Scoring
Assign one score to each non-null field.
- 3: All sentences or content units in the field match the field definition.
- 2: Some sentences or content units in the field do not match the field definition, but the rest do.
- 1: All or nearly all sentences or content units in the field do not match the field definition.

# Rules
- Evaluate field assignment only. Do not judge whether the issue report is factually correct, sufficiently detailed, well written, or useful for solving the task.
"""


BUG_FIELD_DESCRIPTIONS = """
- describe_the_bug: The core problem or the task to address, along with any output or error messages.
- expected_behavior: A clear and concise description of what should have happened instead of the bug.
- reproduction_code: Steps or code to reproduce the bug.
- environment: Runtime environment details, such as OS, language version, and package versions.
- additional_context: Any other context about the problem.
"""


FEATURE_FIELD_DESCRIPTIONS = """
- desired_solution: A clear and concise description of the proposed change or desired behavior.
- problem_motivation: A clear and concise description of the current problem or limitation.
- additional_context: Any other context about the feature request.
"""
\end{lstlisting}

\subsection{Rephrasing Quality Judgment}
\label{app:prompts:judge-rephrasing}

\begin{lstlisting}[style=prompt]
PROMPT_JUDGE_REPHRASING = """
# Role
You are an evaluator assessing whether a GitHub issue, rephrased to resemble real-world requests to AI coding agents such as Claude Code, Codex, or Cursor, preserves previous context.

# Task
Compare `restructured_ps` with `rephrased_ps` as complete texts. Score the following two dimensions independently.

## Information Preservation
Evaluate whether the substantive information in `restructured_ps` remains represented in `rephrased_ps`.
- 3: All information are preserved.
- 2: Some informations are missing, but substantial informations to resolve the task are preserved.
- 1: Substantial information is missing, so the original task cannot be adequately recovered.

## Meaning and Intent Preservation
Evaluate whether `rephrased_ps` preserves the meaning, intent, scope, conditions, relationships of `restructured_ps`.
- 3: Meaning and intent are fully preserved.
- 2: Some details are less precise or somewhat ambiguous, but `rephrased_ps` alone still communicates the same problem and implementation objective.
- 1: Meaning or intent is materially changed, or unsupported content is added in a way that could lead to a different implementation or acceptance target.

# Rules
- Judge each text as a whole, not sentence by sentence.
- Allow paraphrasing, reordering.
- Do not penalize stylistic changes unless they alter meaning, scope.
"""
\end{lstlisting}

\subsection{Linguistic-Dimension Annotation}
\label{app:prompts:annotation}

\begin{lstlisting}[style=prompt]
PROMPT_DIMENSION_CLASSIFICATION = """
# Identity
You are an annotator. Your task is to classify a user prompt sent to an AI coding agent along four linguistic dimensions.

# Input
You will receive a user prompt sent to an AI coding agent (such as Claude Code, Codex, or Cursor).

The user prompt may take various forms. The following are some examples:
- Natural language only.
- Natural language with pasted code or error logs.
- A short fragment.
- A long multi-paragraph passage.

# Task
Assign exactly one category for each of the four dimensions:

1. formality:
   - formal: Uses formal vocabulary and expressions.
   - casual: Uses casual vocabulary and expressions.
   - other: Does not clearly fit either category.

2. sentence_type:
   - declarative: A declarative type.
   - imperative: An imperative type.
   - interrogative: An interrogative type.
   - other: Does not clearly fit any single category.

3. certainty:
   - confident: Expresses certainty.
   - uncertain: Expresses uncertainty.
   - other: Does not clearly fit either category.

4. perspective:
   - first_person: Uses first-person pronouns.
   - non_first_person: Does not use first-person pronouns.
   - other: Does not clearly fit either category.

# Rules
1. Your task is only to assign categories. Do not modify, paraphrase, or extract any content from the user prompt.
2. Assign exactly one category for each of the four dimensions. The four dimensions are independent of each other.
3. Use other only when the user prompt does not clearly fit any other category in each dimension.
4. If the user prompt contains content not authored by the user, classify based on the natural language authored by the user.

# Output
Example output:
{
  "formality": "casual",
  "sentence_type": "imperative",
  "certainty": "confident",
  "perspective": "first_person"
}
"""
\end{lstlisting}

\subsection{Pairwise Style Comparison}
\label{app:prompts:judge-style}

\begin{lstlisting}[style=prompt]
PROMPT_JUDGE_STYLE = """
# Identity
You are an annotator comparing two expressions of the same software-engineering request.

# Task
For each of the four linguistic dimensions below, choose Candidate A, Candidate B, or tie.

1. formality
   Choose the candidate whose differing wording uses a more conversational and everyday lexical register, rather than formal, institutional, or specification-like language.

2. sentence_type
   Choose the candidate that is closer to an imperative.

3. certainty
   Choose the candidate that expresses greater certainty and confidence.

4. perspective
   Choose the candidate that is closer to a non-first-person perspective.

# Rules
1. Judge the four dimensions independently.
2. For every dimension, output exactly A, B, or tie.
3. Evaluate only linguistic expression for the dimension being judged.
4. Do not reward a candidate merely for being longer, more detailed, or more technically complete.
5. Do not evaluate factual correctness, information preservation, task solvability, implementation quality, or solution quality.
6. For formality, compare only lexical choices that differ between the candidates. Ignore sentence type, perspective, length, and unchanged technical content. Output tie when the differing wording shows no meaningful lexical-register difference.
"""
\end{lstlisting}

\end{document}